\PassOptionsToPackage{numbers}{natbib}
\PassOptionsToPackage{table,dvipsnames}{xcolor}
\documentclass{article}

\usepackage[main, final]{neurips_2026}

\usepackage[utf8]{inputenc} %
\usepackage[T1]{fontenc}    %
\usepackage{hyperref}       %
\usepackage{url}            %
\usepackage{booktabs}       %
\usepackage{amsfonts}       %
\usepackage{nicefrac}       %
\usepackage{microtype}      %
\usepackage{xcolor}         %
\usepackage{multirow}
\usepackage{multicol}
\usepackage{array}
\usepackage{makecell}
\usepackage{rotating}
\usepackage{geometry}
\usepackage[table]{xcolor}
\usepackage{amsmath}
\newcommand{\ourmethod}{{Swift Sampling}}

\usepackage[normalem]{ulem}
\newif\ifcrreview
\crreviewfalse
\definecolor{CRAdded}{HTML}{174EBA}
\definecolor{CRDeleted}{HTML}{A33636}
\DeclareRobustCommand{\cradd}[1]{\ifcrreview{\color{CRAdded}#1}\else#1\fi}

\newenvironment{craddedblock}{\begingroup\ifcrreview\color{CRAdded}\fi}{\endgroup}

\usepackage{wasysym}
\usepackage[dvipsnames]{xcolor}
\usepackage[font=footnotesize]{caption}
\usepackage{subcaption}
\usepackage{cleveref}
\usepackage{enumitem}
\usepackage{pifont}
\newcommand{\cmark}{\textcolor{ForestGreen}{\ding{51}}}
\newcommand{\xmark}{\textcolor{red}{\ding{55}}}
\usepackage{wrapfig}
\usepackage{twemojis}
\newcommand{\nogain}[1]{#1\hphantom{ {\tiny ($-0.0$)}}}
\newcommand{\ie}{\textit{i}.\textit{e}., }
\newcommand{\eg}{\textit{e}.\textit{g}., }
\usepackage{ifthen}
\newboolean{comments}
\setboolean{comments}{true} %
\usepackage{todonotes}

\ifthenelse{\boolean{comments}}{
  \setlength{\marginparwidth}{1.5cm}
  \newcommand{\shn}[1]{\todo[color=orange!20, size=\tiny]{Seongheon: #1}}
  \newcommand{\sh}[1]{\textcolor{purple}{Seongheon: #1}}
}{
  \renewcommand{\sh}[1]{}
  \renewcommand{\shn}[1]{}
}
\title{Swift Sampling 
    \raisebox{-0.3\height}{\includegraphics[width=0.73cm]{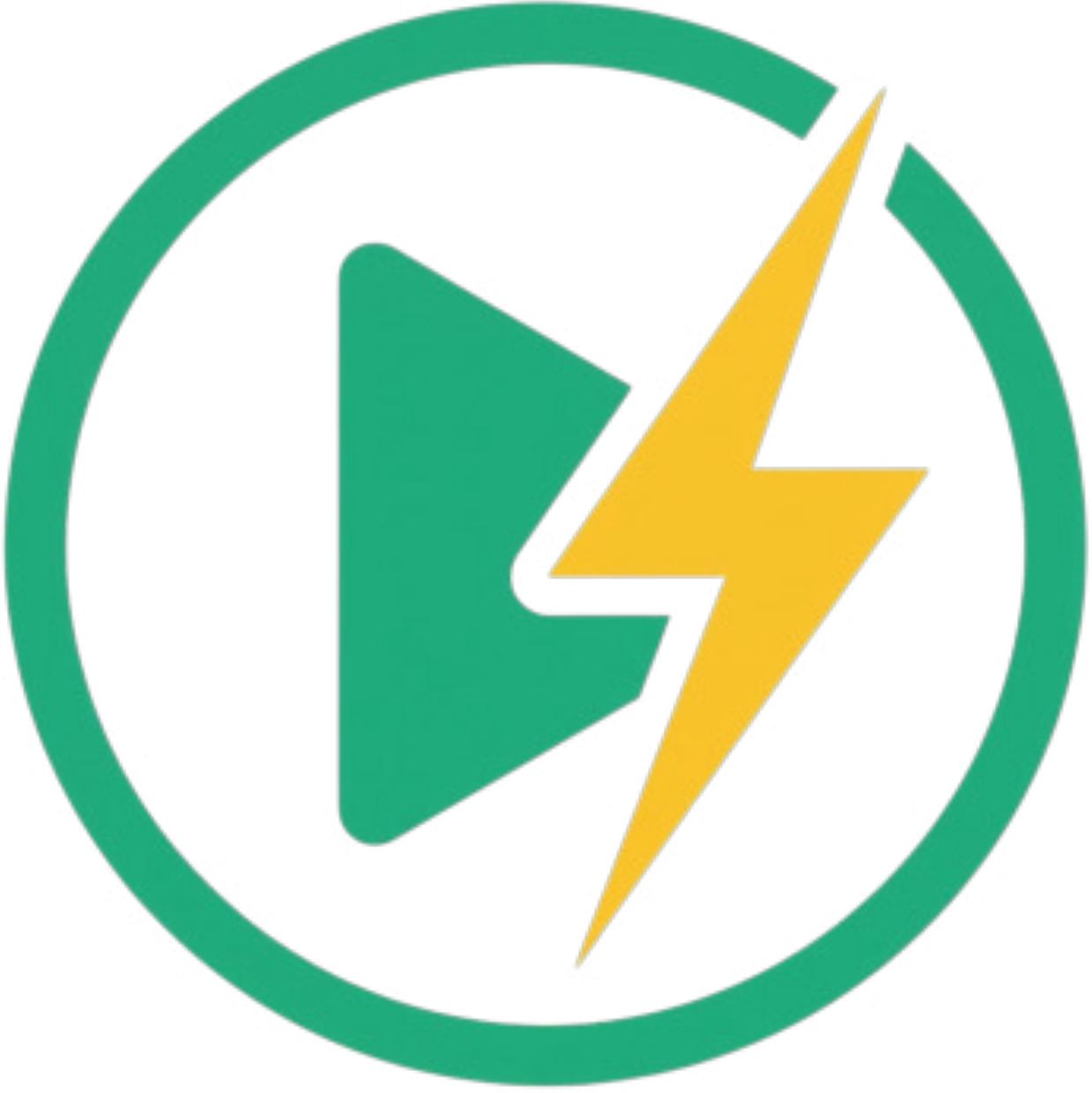}}: 
    Selecting Temporal Surprises via Taylor Series}

\author{%
  \textbf{Dahye Kim}$^{1}$ \hspace{1.5em}
  \textbf{Bhuvan Sachdeva}$^{2}$\thanks{Equal contribution.} \hspace{1.5em}
  \textbf{Karan Uppal}$^{2}$\footnotemark[1] \\[0.4em]
  \textbf{Naman Gupta}$^{2}$\footnotemark[1] \hspace{1.5em}
  \textbf{Vineeth N. Balasubramanian}$^{2}$ \hspace{1.5em}
  \textbf{Deepti Ghadiyaram}$^{1}$ \\[0.7em]
  $^{1}$Boston University \qquad
  $^{2}$Microsoft Research India
}

\begin{document}
\addtocontents{toc}{\protect\setcounter{tocdepth}{-1}}

\maketitle

\begin{figure*}[htbp]
\centering
\vspace{-0.3in}
\includegraphics[width=\textwidth]{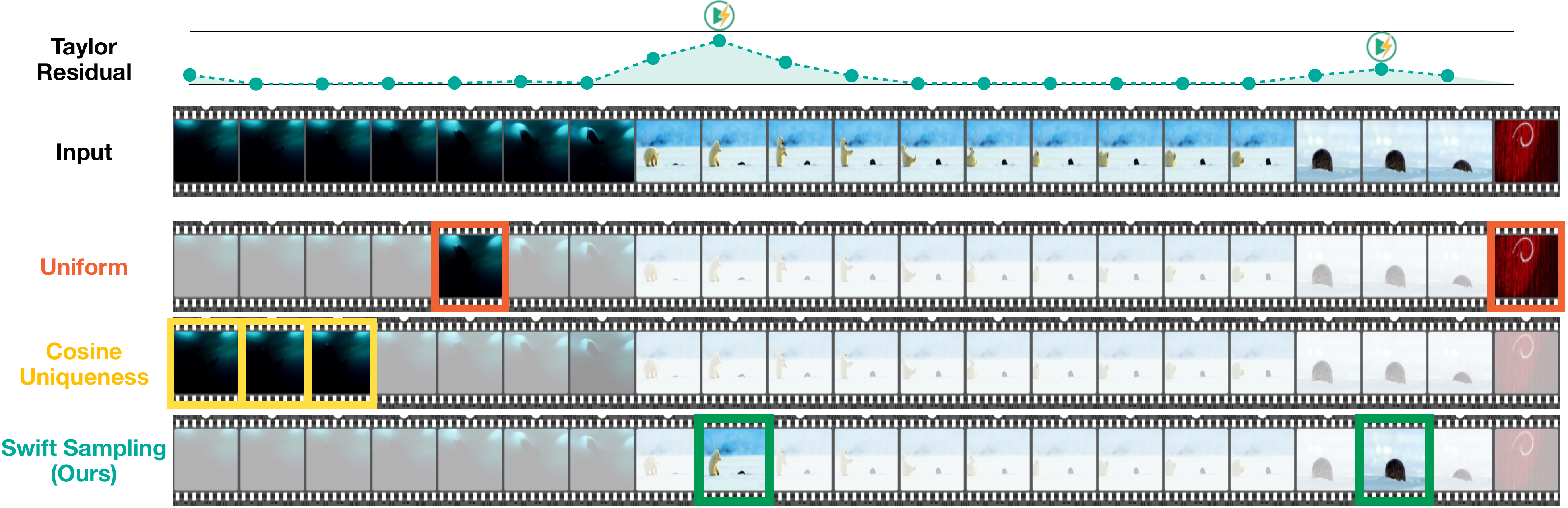}
\caption{\footnotesize {\textbf{Swift Sampling efficiently identifies temporal surprises in videos} by measuring how much a frame deviates from the trajectory predicted by its preceding context. 
Using a Taylor expansion of visual features, we select frames with the largest residuals within their temporal neighborhood as keyframes.
\textit{Top:} Temporal surprise captured using Taylor residual over time.
\textit{Bottom:} input frames and frames selected by Uniform sampling \textcolor{RedOrange}{(orange)}, Cosine Uniqueness~\cite{yuan2025unicomp} \textcolor{YellowOrange}{(yellow)}, and our method \textcolor{ForestGreen}{(green)}. Swift Sampling captures the video’s most informative frames with 30$\times$ less overhead than Cosine Uniqueness, while improving accuracy by up to $+12.5$ points on VQA tasks on long videos with tight frame budgets.}}
\label{fig:teaser}
\end{figure*}
\begin{abstract}
While most frames in long-form video are redundant, the critical information resides in \textit{temporal surprises}: moments where the actual visual features deviate from their predicted evolution. 
Inspired by the human brain’s predictive coding, we introduce \textbf{Swift Sampling}, an elegant, training-free frame selection algorithm that automatically identifies high-information moments in a video. Specifically, we model a video as a differentiable trajectory in the visual latent space and compute the velocity and acceleration of its features. Then, we apply Taylor expansion to project the expected path of subsequent frames. Frames that diverge sharply from this predicted manifold are identified as temporally surprising frames and selected for sampling. Unlike prior training-free methods that rely on auxiliary networks or video-specific hyperparameter tuning, Swift Sampling is incredibly lightweight, adding only $\mathbf{0.02\times}$ additional computational cost over baseline making it $30\times$ cheaper overhead than leading baselines. Across three long-video question answering benchmarks and $10$ different downstream tasks, Swift Sampling outperforms uniform sampling and prior query-agnostic baselines. It is especially powerful for long videos with limited frame budgets improving accuracy by up to $\mathbf{+12.5}$ points.

\end{abstract}
 
\section{Introduction}\label{sec:1}
How does the human brain process the simple sight of a polar bear walking through the snow? Rather than exhaustively processing the continuous visual stream, our visual system is known to operate and revise as a predictive engine: it \textit{anticipates} future states and revises its internal model by calculating the residual errors between its prediction and reality~\cite{rao1999predictive, friston2010free}. As a result, our visual system's computational budget is not wasted on the predictable trajectory of the bear, but is instead reserved for \textit{temporal surprises}, such as the sudden appearance of a seal. This  biological principle inspired seminal
video compression~\cite{cutler1952differential} algorithms and motivates the present
work.

Long-form videos exhibit substantial temporal redundancy: frames evolve slowly
and predictably for extended stretches, punctuated by sparse but
informative transitions.
Yet, most Video Large Language Models (VLMs) still rely on \emph{uniform
sampling} to reduce a video to a fixed frame budget~\cite{zhang2024llava,bai2025qwen3,li2024llava}, not considering temporal structure and treating
redundant frames identically to pivotal ones.
Alternative approaches, such as using optical flow~\cite{teed2020raft}
and pairwise frame-similarity methods~\cite{yuan2025unicomp,scenedetect}, partially
address this, but have their own limitations. First, they require a separate, often external, vision encoder to extract
per-frame representations~\cite{teed2020raft, dinov3,  gmflow, flowformer, zhai2023sigmoid, li2022blip, zhang2024longclip}, nearly doubling the inference cost.
Second, they require careful, video-specific hyperparameter tuning to
define what constitutes a ``significant'' change.
The computational overhead negates the efficiency gains they offer, and
hyperparameter sensitivity can adversely affect downstream task
performance.

Our method is based on a simple observation: long-form video consists of vast, highly predictable intervals interjected with sparse temporal surprises. 
We ask: can we leverage the biologically elegant predictive coding principle to identify these temporal surprises, where a frame’s content diverges from its expected path, without auxiliary models or manual tuning?
To this end, we propose \textbf{Swift Sampling}, a framework that treats the visual latent features of adjacent video frames as points lying on a \emph{locally smooth trajectory} (Fig.~\ref{fig:2}). This makes it amenable to apply a polynomial approximation via Taylor series using higher order derivatives. Given the feature vectors of the $N$ frames preceding the current frame $t$, we construct a Taylor predictor that captures velocity (first order), acceleration (second order), and jerk (third order) of the feature trajectory.
The \textbf{Taylor residual} -- the $\ell_2$ distance between the predicted
and the observed feature -- serves as a principled, per-frame
informativeness score. A small residual indicates a predictable, redundant
frame (\eg a bear's rhythmic walk), while a large residual signals a \emph{temporal surprise}, \ie a moment of genuinely new information (\eg the sudden emergence of seal out of ice). For a given frame budget $K$, we select the $K$ local maxima of the residual sequence, prioritizing the most surprising frame within each local temporal context (Fig.~\ref{fig:teaser}). The sampling rate scales naturally with the video complexity making our approach hyperparameter-light. Crucially, we compute these residuals directly from the \emph{intermediate representations of the VLM's vision encoder} that must be computed anyway during the forward pass.

Our results highlight that the ``temporal surprise'' detection based on Taylor expansion is robust enough to serve as a drop-in replacement for expensive previous methods, bridging the gap between low-level temporal motion and high-level LLM reasoning.
Below, we summarize our contributions:  
\begin{wrapfigure}{r}{0.5\linewidth}
\vspace{1em}
\centering
\includegraphics[width=\linewidth]{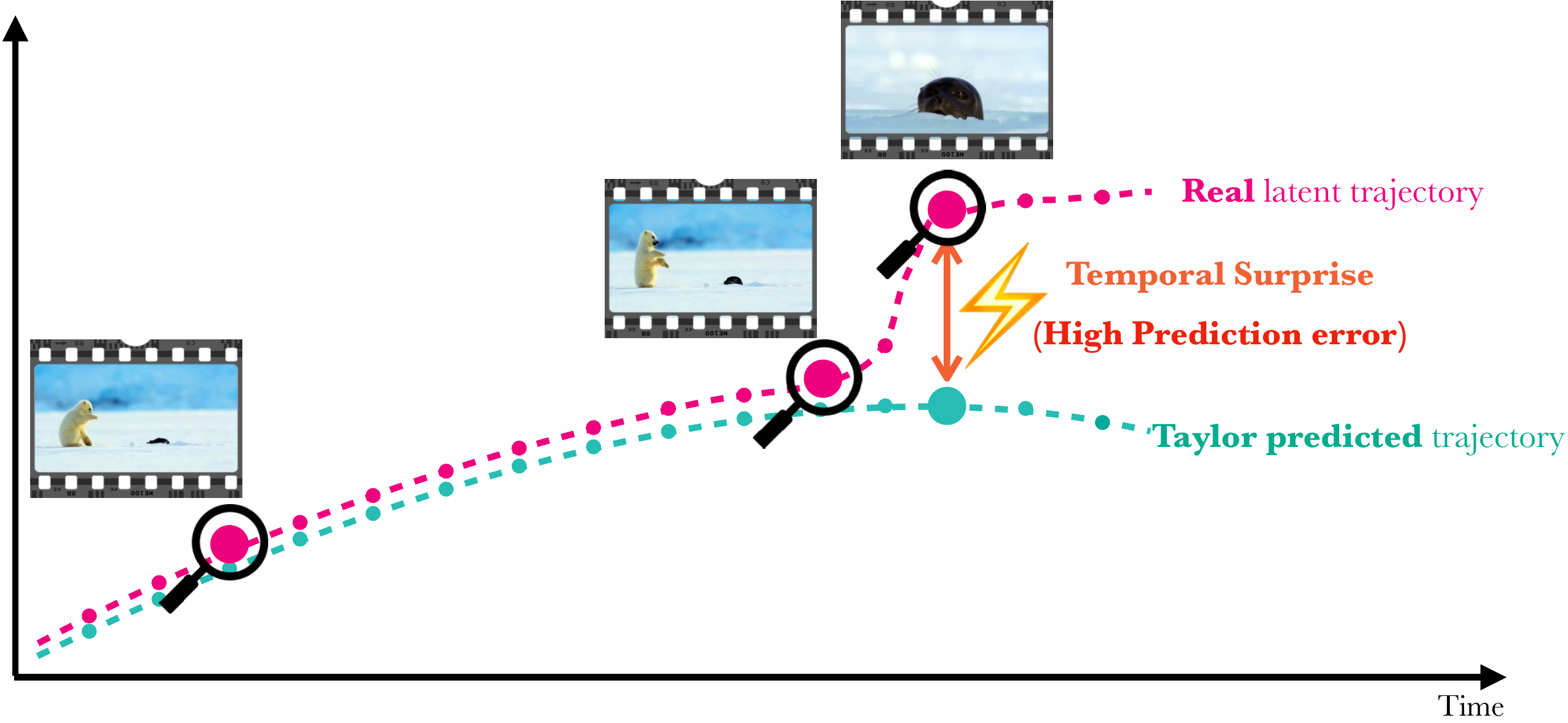}
\caption{\footnotesize Each frame is represented on the \textcolor{RubineRed}{latent feature trajectory}, where we apply Taylor expansion over preceding frames to \textcolor{JungleGreen}{predict the next frame} feature. The residual between the prediction and the actual feature measures how much the trajectory deviates from a smooth continuation. Frames with large residuals correspond to \emph{temporal surprises}, \eg seal suddenly emerging from the ice, which {\ourmethod} effectively captures.} 
\label{fig:2}
\vspace{-0.2in}
\end{wrapfigure}

\vspace{-0.2in}
\begin{itemize}[leftmargin=*,noitemsep]
    \item We propose \textbf{Swift Sampling}, a
    training-free frame selection algorithm that is inspired by predictive coding and scores frames via their Taylor series residual in the VLM's
    latent space, with no auxiliary model or any video-specific tuning making it  hyperparameter-light and efficient.
    \item {\ourmethod} achieves \textbf{state-of-the-art performance} over uniform sampling and several prior training-free methods across different VLM backbones on video question answering, token compression, and over ten other reasoning tasks across diverse video lengths.
    \item We provide a \textbf{systematic analysis} of the design choices of {\ourmethod}, yielding critical insights into the relationship between latent temporal dynamics and frame selection. 
\end{itemize}

\section{Related Work}\label{sec:2}
\noindent\textbf{Video large language models and long video understanding.}
Video large language models have achieved impressive results on short-form video understanding~\cite{zhang2024llava, bai2025qwen3, lin2024video,li2025videochat,jin2024chat,cheng2024videollama,liu2024llavanext,fei2024video,wang2024qwen2,chen2024expanding}, but processing long videos remains challenging due to the large number of input frames.
To better handle long-form inputs, prior works improve temporal modeling~\cite{zhang2024llava, li2025videochat,cheng2024videollama}, multimodal fusion~\cite{li2024llava, lin2024video,jin2024chat,fei2024video}, and multi-scale encoding~\cite{bai2025qwen3, liu2024llavanext, wang2024qwen2, chen2024expanding,  xu2025slowfast,team2025kwai}; others explicitly target long videos through context-length extension~\cite{team2025kwai, chen2024longvila,zhang2024long}, temporal token compression~\cite{fei2024video, shen2024longvu,cheng2025scaling}, or KV-cache sparsification~\cite{shu2025video}.
Despite these advances, most approaches still rely on uniform sampling to reduce raw videos to a fixed number of frames, overlooking redundancy among sampled frames. We focus on this preprocessing stage, selecting non-redundant frames to make better use of the limited frame budget, which is orthogonal and complementary to these model-level improvements.

\noindent\textbf{Frame selection for long video understanding.}
Frame selection methods for long-video understanding have been actively explored along two directions: training-based and training-free approaches.
Training-based methods learn to select frames through end-to-end optimization with downstream task losses~\cite{buch2022revisiting,buch2025flexible}, frame-candidate ranking~\cite{yu2024frame}, pseudo-label supervision from vision-language models~\cite{hu2025m}, reinforcement or self-learning~\cite{xu2025viarl,lee2025refocus,yu2023self, yang2025cambrian}, and supervised keyframe annotations~\cite{yao2025generative,ghazanfari2025chain}. Although effective, these methods often require additional training or adaptation for each VLM~\cite{buch2025flexible,yu2024frame,hu2025m}, which is expensive and limits practical deployment.
To avoid this limitation, training-free frame selection methods have been preferred.
Query-aware methods have been heavily explored~\cite{tang2025adaptive,sun2025frames,zhang2025q,sun2025mdp3,arnab2025temporal,hu2025cos,zhu2025focus,liu2025bolt,zhang2025adard}, which select frames based on text-visual similarity with the language query.
Query-agnostic methods~\cite{li2026maxinfo} select frames solely from visual features without access to the query.
However, both categories typically require encoding all candidate frames with a separate vision encoder to compute frame-level representations, which can nearly double inference cost. By contrast, {\ourmethod} avoids the need for an auxiliary model by leveraging the VLM’s own vision encoder, thereby incurring negligible computational overhead.

\noindent \textbf{Tokenization-based} approaches such as ElasticTok~\cite{elastictok}, EVATok~\cite{evatok}, AdapTok~\cite{adaptok}, SURGE~\cite{tangsurge} and InfoTok~\cite{infotok}, dynamically adjust the number of tokens according to video content complexity. Similarly, methods such as ToMe~\cite{tome} and PruneVid~\cite{prunevid} focus on efficiency by merging spatially or temporally redundant tokens. In contrast, {\ourmethod} first identifies the most informative frames to retain prior to tokenization. By filtering redundant frames at the input level, {\ourmethod} offers a complementary layer of efficiency that can be combined with token-level compression strategies. 

\noindent\textbf{Taylor series for video understanding.}
The Taylor series approximates a function at a given point using its derivatives, decomposing local behavior into zeroth-order (value), first-order (velocity), second-order (acceleration), and higher-order terms.
This predictive structure has been used in video understanding and generation. %
Taylor Video~\cite{wang2024taylor} sums higher-order Taylor residuals into a dedicated motion representation that replaces or complements RGB frames as input to action classifiers; ViDiDi~\cite{chen2024unfolding} uses temporal derivatives as additional views for self-supervised video representation learning.
More recently, TaylorSeer~\cite{liu2025reusing} and SCOPE~\cite{cui2026not} use Taylor prediction to estimate future features across diffusion denoising steps, skipping recomputation when the prediction is reliable.
These works use Taylor terms primarily to construct new representations or accelerate generation. In contrast, we use the magnitude of the Taylor residual as a frame-level informativeness score: frames whose features deviate strongly from their predicted trajectory are treated as informative and selected as keyframes. While Taylor expansions have been used before, we are not aware of any prior works that use them as a training-free, query-agnostic frame selector for very long videos.
\section{\ourmethod: Our Approach}\label{sec:3}
Given a video with $T$ frames and a target budget of $K \le T$, our objective is to select the $K$ most informative frames for a downstream video model. To achieve this, we propose {\ourmethod}, a selection strategy grounded in the Taylor series expansion of latent visual features. %
Sec.~\ref{sec:3.1} introduces the Taylor predictor for
latent feature sequences, and Sec.~\ref{sec:3.2} formalizes the Taylor residual as a principled informativeness score and presents the full selection algorithm. %
\subsection{Background: Taylor Series Expansion for Sequence Prediction}
\label{sec:3.1}
Let $x$ be a smooth scalar-valued function of time, let $t_0$ denote the
current timestep and let $x^{(n)}(t_0)$ denote the $n$-th derivative of
$x$ at $t_0$. The Taylor series predicts $x$ at a future time
$t_0 + \Delta t$ from higher order derivatives of $x$, defined as follows:
\begin{equation}
\small
x(t_0 + \Delta t) \;=\; \sum_{n=0}^{\infty} \frac{\Delta t^n}{n!}\, x^{(n)}(t_0) \;=\; x(t_0) + \Delta t \cdot x^{(1)}(t_0) + \tfrac{\Delta t^2}{2!} x^{(2)}(t_0) + \cdots.
\label{eq:taylor_continuous}
\end{equation}
In practice, $x$ is observed only at discrete timesteps, so derivatives
must be approximated by \emph{backward finite differences}~\cite{leveque2007finite}. The first-order derivative is approximated as the difference between the two most recent samples,
\begin{equation}
\small
x^{(1)}(t_0) \;\approx\; \frac{x(t_0) - x(t_0 - \Delta t)}{\Delta t},
\label{eq:fd_first}
\end{equation}
and the second-order derivative as the difference of two consecutive first-order differences,
\begin{equation}
\small
x^{(2)}(t_0) \;\approx\; \frac{x^{(1)}(t_0) - x^{(1)}(t_0 - \Delta t)}{\Delta t} \;=\; \frac{x(t_0) - 2 x(t_0 - \Delta t) + x(t_0 - 2\Delta t)}{\Delta t^2}.
\label{eq:fd_second}
\end{equation}

In general, the $n$-th order approximation is a linear combination of current and $n$ preceding frames (thus $n+1$ total frames).
This is derived by applying the difference operator $n$ times to the sequence $x(t_0), x(t_0 - \Delta t), \dots, x(t_0 - n\Delta t)$, with weights determined by binomial coefficients.
\begin{equation}
\small
  x^{(n)}(t_0) \approx
    \frac{1}{\Delta t^n}
    \sum_{k=0}^{n}(-1)^k \binom{n}{k} x(t_0 - k\,\Delta t).
  \label{eq:fd_general}
\end{equation}
Substituting these estimates into Eq.~\eqref{eq:taylor_continuous} yields a \textit{closed-form linear combination of preceding samples}, enabling efficient prediction of $x(t_0 + \Delta t)$ directly from observations.

\subsection{Taylor Residual as an Informativeness Signal}
\label{sec:3.2}
\noindent\textbf{Using the Taylor residual.} Let $f_t \in \mathbb{R}^d$ denote the visual feature vector extracted from the video frame at time $t$ and let $f^{(n)}_{t-1}$ denote the $n$-th order derivative of the visual feature trajectory
at time $t-1$. A natural criterion for frame informativeness under a fixed budget is
\emph{temporal surprise}: frame (feature) $f_t$ is informative if its content is not
predictable from the preceding context $f_1,\dots,f_{t-1}$.  Predictive
coding theory formalizes this intuition by equating informativeness with
prediction error, \ie the discrepancy between the observed signal and the best prediction derived from prior context~\cite{rao1999predictive}. 

For a latent feature trajectory that evolves smoothly in time, the natural local
predictor is the Taylor expansion $\hat{f}_t \;=\; f_{t-1} + f'_{t-1}\,\Delta t + \tfrac{1}{2}\,f''_{t-1}\,(\Delta t)^{2} + \cdots$, 
which extrapolates the trajectory under the assumption of locally polynomial dynamics. 

As noted in Eq.~\ref{eq:fd_general}, the $n$-th order derivative can be approximated using backward finite-differences from the sequence of preceding features, \ie $\{f_{t-1}, \ldots, f_{t-1-n}\}$. Assuming uniform temporal spacing ($\Delta t = 1$) and truncating Eq.~\eqref{eq:taylor_continuous} at order $N$, following prior works~\cite{leveque2007finite, baeza2017approximate,pourheydari2021taylorswiftnet,  liu2025reusing}, we define the Taylor predictor of $f_t$ based on its $N+1$ predecessors as: 
\begin{equation}
\small
\hat{f}_t^{(N)} \;=\; f_{t-1} + f^{(1)}_{t-1} + \tfrac{1}{2!}\,f^{(2)}_{t-1} + \cdots + \tfrac{1}{N!}\,f^{(N)}_{t-1}
\label{eq:taylor_frame}
\end{equation}

Now, temporal surprise or Taylor residual at frame $t$ is the magnitude of the prediction error:
\begin{equation}
\small
r_t \;=\; \left\| f_t - \hat{f}_t^{(N)} \right\|_2 .
\label{eq:taylor_residual}
\end{equation}

While the Taylor predictor captures the trajectory's local kinematic structure such as velocity, acceleration, jerk, etc.,  $r_t$ isolates the \emph{surprise}, the component of $f_t$ not explained by smooth extrapolation. Concretely, frames with a large residual $r_t$ deviate sharply from the predicted trajectory, indicating high information content relative to their temporal context. Conversely, frames with a small $r_t$ closely adhere to the predicted path and are considered redundant. Consequently, the Taylor residual sequence $\{r_t\}$ provides a principled, per-frame informativeness signal across the candidate pool.

\noindent\textbf{Information-theoretic motivation:} From a statistical perspective, under an isotropic Gaussian model for innovation (novel information)
$f_t = \hat{f}_t^{(N)} + \epsilon_t,\;
 \epsilon_t \sim \mathcal{N}(\mathbf{0},\sigma^2 I)$, following \cite[Ch.~8, Thm~8.4.1]{cover2006elements}, the Shannon self-information (surprise) of frame (feature)~$f_t$ given its context can be written as:
\begin{equation}\label{eq:surprise}
\small
  -\log p\bigl(f_t \mid f_1,\dots,f_{t-1}\bigr)
  \;=\;
  \frac{1}{2\sigma^2} r_t^2
  \;+\;\mathrm{const}\,,
\end{equation}
which is monotonically increasing in the Taylor-residual magnitude $r_t$.  While this Gaussian model is an idealization (we do not claim that the vision encoder's projections are Gaussian), it motivates our use of Taylor residual as a tractable surrogate for informativeness. We note that this interpretation is consistent with classical filtering formulations, where larger innovations induce larger posterior corrections (\eg~\cite{bishop2006prml}).

\noindent{\textbf{Local Maxima Selection:}} 
A key subtlety is that $r_t$ is computed relative to its
predecessors, so its absolute scale depends on the local dynamics of the
trajectory: a slow, uniform scene yields consistently low residuals, while
a fast-moving segment produces uniformly high values.
Consequently, selecting the global top-$K$ residuals would concentrate all
keyframes within a few high-motion bursts, leaving subtler but critical
events entirely unrepresented. We therefore select the \emph{local maxima} of $\{r_t\}$, identifying the
most surprising frame within each local temporal context, regardless of
absolute magnitude.
Formally, let $M<T$ denote the number of detected local maxima, which vary with the video content.
Formally, we define the set of local maxima $\mathcal{P} = \{p_1, \ldots, p_M\}$, indexed in increasing temporal order as: $\mathcal{P} \;=\; \bigl\{\, i \,:\, r_i > r_{i-1} \;\text{and}\; r_i > r_{i+1} \,\bigr\}$. From this candidate set $\mathcal{P}$, we select the $K$ elements with the largest residuals to serve as the final frames. In cases where the video is highly static ($M < K$), the remaining $K - M$ slots are filled using the highest-residual frames from the pool of non-maxima, $\{1, \ldots, T\} \setminus \mathcal{P}$. We demonstrate in Sec.~\ref{sec:4} that this hierarchical selection strategy prioritizes the most significant surprises relative to their immediate context.

\section{Experiments}\label{sec:4}
\begin{figure*}[t]
\centering
\includegraphics[width=\textwidth]{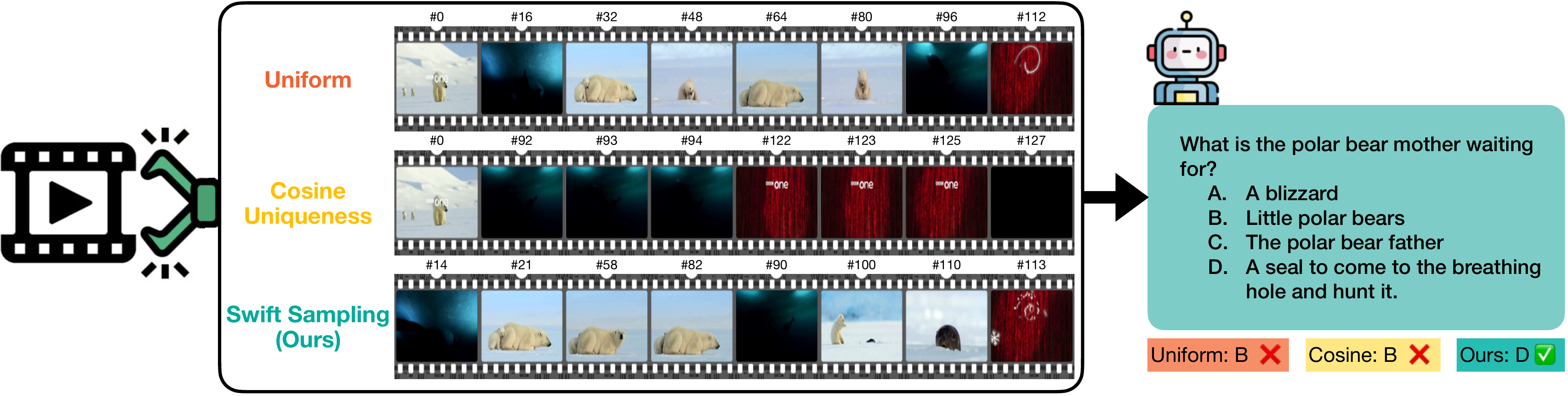}\vspace{-0.1in}
\caption{\footnotesize \textbf{Qualitative comparison of frame selection} on a sample video from VideoMME dataset, given a budget of $8$ frames out of $128$. The correct answer (a seal at the breathing hole) requires temporal coverage of multiple events. Uniform sampling is redundant, capturing the polar bear while missing the seal entirely.
Cosine Uniqueness~\cite{yuan2025unicomp} favors visual outliers like title cards and underwater shots that are task irrelevant and fail to provide relevant information. By contrast, Swift Sampling captures the temporal surprise of the seal's appearance thus providing critical evidence for correct reasoning.}
\label{fig:qual_example}
\end{figure*}
 \begin{table*}[t]
\centering
\caption{\footnotesize \textbf{VQA accuracy across different video durations} on Video-MME, LongVideoBench (LVB), and MLVU benchmark. The \textbf{Query-agnostic} column indicates if the method selects frames without using the query (\cmark: query-agnostic, \xmark: query-aware). The \textbf{FLOPs} column reports inference cost \textit{relative} to uniform sampling on the same backbone ($1.00\times$ = no overhead beyond the base VLMs forward pass). Within each (backbone, query-type) block: \textbf{bold} = best, \underline{underline} = second best.}\vspace{-7pt}
\label{tab:main_results}
\resizebox{\textwidth}{!}{%
\begin{tabular}{l c c c|
  >{\columncolor{RedOrange!4}}c >{\columncolor{RedOrange!8}}c >{\columncolor{RedOrange!14}}c >{\columncolor{RedOrange!22}}c |  
  >{\columncolor{Goldenrod!4}}c >{\columncolor{Goldenrod!8}}c >{\columncolor{Goldenrod!14}}c >{\columncolor{Goldenrod!22}}c |
  >{\columncolor{JungleGreen!4}}c >{\columncolor{JungleGreen!8}}c >{\columncolor{JungleGreen!14}}c >{\columncolor{JungleGreen!22}}c}
\toprule
\multirow{2}{*}{Method} & \multirow{2}{*}{\shortstack{Size /\\FLOPs}} & \multirow{2}{*}{\shortstack{Query-\\agnostic}} & \multirow{2}{*}{\# Frames} & \multicolumn{4}{c}{Video-MME} & \multicolumn{4}{c}{LVB} & \multicolumn{4}{c}{MLVU} \\
\cmidrule(lr){5-8} \cmidrule(lr){9-12} \cmidrule(lr){13-16}
& & & & Short & Medium & Long & Overall & $\geq$10m & $\geq$20m & $\geq$30m & Overall & $\geq$10m & $\geq$15m & $\geq$30m & Overall \\
\midrule

\multicolumn{16}{l}{\textbf{Pretrained VLLM w/ Uniform Sampling:}} \\

\quad VideoChat2~\cite{li2025videochat}      & 7B & \cmark & 32    & 36.7 & 31.7 & 28.6 & 32.3 & 22.4    & 19.6    & 15.7    & 22.6 & 46.1    & 41.6    & 43.8    & 50.6 \\
\quad VideoLLaMA3~\cite{zhang2025videollama} & 7B & \cmark & 32 & 77.2 & 61.7 & 53.4 & 64.1 & 49.1  & 51.1    & 52.8    & 58.0    & 56.5    & 53.7    & 45.8 & 57.2   \\

\quad LongVA~\cite{zhang2024long}            & 7B & \cmark & 32  & 63.2 & 50.7 & 45.0 & 53.0 & 46.5    & 46.7    & 46.3    & 52.6    & 56.5    & 50.3    & 54.2    & 57.4    \\

\quad Qwen2.5-VL~\cite{bai2025qwen25vl}          & 7B & \cmark & 32    & 72.6    & 61.2    & 50.2    & 61.3 & 50.9    & 54.3    & 59.3    & 58.8 & 59.6    & 49.0    & 54.2    & 60.3 \\

\quad InternVL3~\cite{zhu2025internvl3}      & 8B & \cmark & 32      & 75.3    & 64.4    & 54.3    & 64.7 & 48.9    & 50.0    & 49.1    & 58.9 & 68.9    & 61.1    & 64.6    & 70.1 \\
\midrule

\multicolumn{16}{l}{\textbf{\colorbox{Periwinkle!22}{Training-based Frame Selection:}}} \\

\quad Frame-Voyager~\cite{yu2024frame}       & 7B   & \xmark & 8/32   & 67.3 & 56.3 & 48.9 & 57.5 & -    & -    & -    & -    & -    & -    & -    & 65.6 \\

\quad Hu et al.~\cite{hu2025m}               & 8.5B & \xmark & 128/32 & 69.6 & 54.1 & 51.9 & 58.7 & -    & -    & -    & -    & -    & -    & -    & -    \\

\quad GenS~\cite{yao2025generative}          & 7B   & \xmark & 54/32  & -    & -    & -    & -    & -    & -    & -    & 58.7 & -    & -    & -    & 64.8 \\
\midrule
\multicolumn{16}{l}{\textbf{\colorbox{RubineRed!22}{Training-free Frame Selection:}}} \\
\multicolumn{16}{l}{\textit{LLaVA-OneVision~\cite{li2024llava}}} \\
\quad + \textit{Uniform}                                       & 1.00$\times$ & \cmark & 128$\to$32 & \underline{69.9} & \underline{56.4} & \underline{48.8} & \underline{58.3} & 45.2 & \underline{47.5} & \underline{48.1} & \underline{55.3} & 61.4 & 54.4 & \underline{50.0} & 64.7 \\
\quad + \textit{Cosine Uniqueness}~\cite{yuan2025unicomp}      & 1.60$\times$ & \cmark & 128$\to$32 & 65.3 & 54.7 & 47.0 & 55.7 & \underline{47.0} & 47.1 & 46.3 & 52.5 & \textbf{63.6} & \textbf{61.1} & 47.9 & \underline{65.4} \\
\quad + \textit{\ourmethod\ (Ours)}                            & \hspace{6pt} 1.02$\times$ \twemoji{rocket}& \cmark & 128$\to$32 & \textbf{71.0} & \textbf{56.9} & \textbf{49.2} & \textbf{59.0} & \textbf{51.6} & \textbf{54.3} & \textbf{50.9} & \textbf{57.9} & \underline{62.2} & \underline{58.2} & \textbf{54.2} & \textbf{65.6} \\
\midrule
\quad + \textit{AKS}~\cite{tang2025adaptive}                   & 1.53$\times$ & \xmark & 128$\to$32 & 66.7 & \textbf{56.4} & \textbf{48.8} & 57.3 & \underline{54.8} & 52.7 & \textbf{52.7} & 58.0 & \textbf{65.7} & \underline{59.1} & \textbf{56.2} & 68.6 \\
\quad + \textit{AKS}~\cite{tang2025adaptive}                   & 2.06$\times$ & \xmark & 256$\to$32 & 62.0 & 52.8 & 48.4 & 54.4 & \textbf{54.9} & \textbf{55.1} & 49.1 & \underline{58.7} & \underline{65.0} & \textbf{61.1} & \textbf{56.2} & \underline{69.4} \\
\quad + \textit{AKS+\ourmethod\ (128$\to$96)}                  & 1.43$\times$ & \xmark & 128$\to$32 & \textbf{67.4} & \underline{56.1} & \underline{48.7} & \textbf{57.4} & 54.1 & 52.1 & \underline{52.1} & 58.6 & 64.8 & 57.7 & \underline{50.0} & \underline{69.4} \\
\quad + \textit{AKS+\ourmethod\ (256$\to$128)}                 & 1.59$\times$ & \xmark & 256$\to$32 & \underline{67.0} & 54.7 & 47.8 & 56.5 & 54.0 & \underline{52.5} & \underline{52.5} & \textbf{58.9} & 64.6 & \underline{59.1} & \underline{50.0} & \textbf{69.9} \\
\midrule
\multicolumn{16}{l}{\textit{LLaVA-Video~\cite{zhang2024llava}}} \\
\quad + \textit{Uniform}                                       & 1.00$\times$ & \cmark & 128$\to$32 & \underline{74.0} & \underline{59.0} & \underline{51.3} & \underline{61.4} & \underline{50.7} & \underline{53.3} & \underline{54.6} & \underline{56.8} & 56.7 & 54.4 & 50.0 & 64.2 \\
\quad + \textit{Cosine Uniqueness}~\cite{yuan2025unicomp}      & 1.60$\times$ & \cmark & 128$\to$32 & 69.2 & 54.4 & 50.1 & 57.9 & \underline{50.7} & 52.5 & \underline{54.6} & 56.5 & \textbf{61.2} & \textbf{57.0} & 50.0 & \underline{66.5} \\
\quad + \textit{\ourmethod\ (Ours)}                            & \hspace{6pt} 1.02$\times$ \twemoji{rocket} & \cmark & 128$\to$32 & \textbf{74.9} & \textbf{59.1} & \textbf{51.6} & \textbf{61.9} & \textbf{52.1} & \textbf{56.2} & \textbf{57.4} & \textbf{58.6} & \underline{60.0} & \underline{55.0} & \textbf{52.1} & \textbf{67.2} \\
\bottomrule
\end{tabular}
}
\end{table*} 
\begin{table*}[t]
\centering
\caption{\footnotesize \textbf{Comparison with additional query-agnostic baselines on LLaVA-OneVision.} All methods select $K{=}32$ frames from a pool of $128$ candidates. \textbf{FLOPs} report inference cost relative to uniform sampling. \textbf{Bold}: best, \underline{underline}: second best.}
\vspace{-0.1in}
\label{tab:query_agnostic_analysis}
\resizebox{\textwidth}{!}{%
\begin{tabular}{l c c|
  >{\columncolor{RedOrange!4}}c >{\columncolor{RedOrange!8}}c >{\columncolor{RedOrange!14}}c >{\columncolor{RedOrange!22}}c |
  >{\columncolor{Goldenrod!4}}c >{\columncolor{Goldenrod!8}}c >{\columncolor{Goldenrod!14}}c >{\columncolor{Goldenrod!22}}c |
  >{\columncolor{JungleGreen!4}}c >{\columncolor{JungleGreen!8}}c >{\columncolor{JungleGreen!14}}c >{\columncolor{JungleGreen!22}}c}
\toprule
\multirow{2}{*}{Method} & \multirow{2}{*}{FLOPs} & \multirow{2}{*}{\# Frames} & \multicolumn{4}{c}{Video-MME} & \multicolumn{4}{c}{LVB} & \multicolumn{4}{c}{MLVU} \\
\cmidrule(lr){4-7} \cmidrule(lr){8-11} \cmidrule(lr){12-15}
& & & Short & Medium & Long & Overall & $\geq$10m & $\geq$20m & $\geq$30m & Overall & $\geq$10m & $\geq$15m & $\geq$30m & Overall \\
\midrule
Uniform                                       & 1.00$\times$ & 128$\to$32 & 69.9 & 56.4 & \underline{48.8} & 58.3 & 45.2 & 47.5 & 48.1 & 55.3 & 61.4 & 54.4 & 50.0 & 64.7 \\
Cosine Uniqueness~\cite{yuan2025unicomp}      & 1.60$\times$ & 128$\to$32 & 65.3 & 54.7 & 47.0 & 55.7 & 47.0 & 47.1 & 46.3 & 52.5 & \underline{63.6} & \textbf{61.1} & 47.9 & 65.4 \\
Frame difference                              & 1.00$\times$ & 128$\to$32 & 67.4 & 53.3 & 48.3 & 56.4 & 46.3 & 49.3 & \textbf{51.9} & 53.5 & 58.5 & 53.7 & 45.8 & 64.6 \\
Iframe                                        & 1.00$\times$ & 128$\to$32 & 67.4 & 54.9 & 48.7 & 57.0 & \underline{52.0} & 49.8 & 49.8 & 57.1 & 60.6 & 54.4 & \underline{52.1} & 63.8 \\
Pframe                                        & 1.00$\times$ & 128$\to$32 & 66.9 & 55.1 & 48.2 & 56.7 & 51.9 & 49.1 & 49.1 & 56.5 & 60.8 & 54.4 & 50.0 & 64.1 \\
Optical Flow~\cite{teed2020raft}                      & 1.07$\times$ & 128$\to$32 & 68.6 & 53.0 & 48.0 & 56.5 & \textbf{52.4} & 50.7 & 50.7 & 56.9 & 60.0 & 56.4 & \underline{52.1} & 62.9 \\
DySeg (adapted)~\cite{shen2025fastvid}        & 1.79$\times$ & 128$\to$32 & 69.6 & 53.7 & 48.4 & 57.2 & 46.0 & 48.2 & \textbf{51.9} & 52.9 & 49.6 & 47.7 & 48.3 & 63.1 \\
MaxInfo~\cite{li2026maxinfo}                  & 1.79$\times$ & 128$\to$32 & \textbf{71.1} & \textbf{57.2} & \underline{48.8} & \underline{58.9} & 51.4 & \underline{50.8} & 50.0 & \underline{57.8} & \textbf{63.0} & \textbf{59.1} & 51.1 & \textbf{66.5} \\ 
\midrule
\ourmethod\ (Ours)                            & \hspace{6pt} 1.02$\times$ \twemoji{rocket} & 128$\to$32 & \underline{71.0} & \underline{56.9} & \textbf{49.2} & \textbf{59.0} & 51.6 & \textbf{54.3} & \underline{50.9} & \textbf{57.9} & \underline{62.2} & \underline{58.2} & \textbf{54.2} & \underline{65.6} \\
\bottomrule
\end{tabular}
} 
\end{table*}

\subsection{Experimental Settings}\label{sec:4_setup}
\noindent\textbf{Benchmarks.}
We evaluate on three well-known long video benchmarks:
Video-MME~\cite{fu2025video},
MLVU~\cite{zhou2025mlvu}, and LongVideoBench (LVB)~\cite{wu2024longvideobench}. Each benchmark focuses specifically on Visual Question Answering (VQA) as the primary downstream reasoning task.
To measure how frame selection quality scales with video duration, we report accuracy across various temporal subsets, ranging from short clips to videos exceeding 30 minutes.

\noindent\textbf{Baselines.} 
We compare {\ourmethod} against
\begin{itemize}[nosep]
    \item \colorbox{Periwinkle!22}{training-based} supervised methods~\cite{yu2024frame, hu2025m, yao2025generative} designed specifically for frame selection,
    \item \colorbox{RubineRed!22}{training-free} query-aware methods (marked as \xmark\ in Table~\ref{tab:main_results}), that utilize the input question to identify relevant frames during inference~\cite{tang2025adaptive},
    \item \colorbox{RubineRed!22}{training-free} query-agnostic methods (marked as \cmark\ in Table~\ref{tab:main_results}) that select frames based purely on video content, including MaxInfo~\cite{li2026maxinfo} and the following baselines: 1) \textbf{Cosine Uniqueness}~\cite{yuan2025unicomp}, which selects the top-$K$ most unique frames based on inter-frame cosine similarity; 2) \textbf{Frame Difference}, which selects the top-$K$ frames with the largest adjacent-frame feature differences; 3) \textbf{I-Frame} and 4) \textbf{P-Frame}, which select the top-$K$ frames ranked by I-frame or P-frame packet sizes from the video codec, respectively;
5) \textbf{Optical Flow Based}, which uses the pretrained optical flow estimator RAFT~\cite{teed2020raft} 
to pick top-K frames with the highest mean flow magnitude  
6) \textbf{DySeg}~\cite{shen2025fastvid}, originally proposed for segment grouping, which we adapt for frame selection. Details in the appendix.
\end{itemize}

\noindent\textbf{Implementation details.}
We use two representative VLM backbones for our experiments: LLaVA-OneVision~\cite{li2024llava} and LLaVA-Video~\cite{zhang2024llava}. 
Given a video, we uniformly sample 128 candidate frames and select $K=32$ frames for downstream tasks. 
We extract frame representations from the key projections of the vision encoder's first transformer layer ($\ell=0$) to compute the Taylor residuals. The spatial tokens are mean-pooled to obtain the per-frame feature $f_t$ used in Eq.~\ref{eq:taylor_residual}. Throughout all experiments, we fix the Taylor expansion order to $N=3$. For all baselines, we strictly adhere to their publicly available implementations. More details in Appendix.

\subsection{Main Results}
Results are summarized in Tables~\ref{tab:main_results} and ~\ref{tab:query_agnostic_analysis}. 
{\ourmethod} brings consistent gains over uniform sampling across all two backbones, with particularly strong improvements on long-duration videos. 
Using \textit{LLaVA-OneVision} as a backbone, on the LVB dataset, the overall accuracy improves from $55.3$ to $57.9$ ($+2.6$); on MLVU dataset, the overall jumps from $64.7$ to $65.6$ ($+0.9$). 
The gains are more pronounced on longer videos: $+6.8$ points on LVB videos longer than $20$ minutes ($47.5 \to 54.3$), $+6.4$ points on LVB videos longer than $10$ minutes ($45.2 \to 51.6$), and $+4.2$ points on MLVU videos longer than $30$ minutes ($50.0 \to 54.2$).
On \textit{LLaVA-Video}, we observe similar trends: MLVU overall improves from $64.2$ to $67.2$ ($+3.0$), and LVB videos longer than $20$ minutes from $53.3$ to $56.2$ ($+2.9$). 
As shown in Fig.~\ref{fig:qual_example}, these improvements stem from our method’s ability to capture pivotal ``surprise'' events that standard baselines overlook.

On a broader comparison with query-agnostic baselines (\cmark\ in Table~\ref{tab:main_results} and the entirety of  Table~\ref{tab:query_agnostic_analysis}), our method remains highly competitive while operating at a negligible $1.02\times$ inference cost. This efficiency stems from reusing the target Video LLM’s existing vision encoder -- specifically the first few layers -- to compute frame representations. This adds only $0.02\times$ to the total inference cost compared to the vanilla model. 
By contrast, existing \colorbox{RubineRed!22}{training-free} methods~\cite{yuan2025unicomp, tang2025adaptive, shen2025fastvid, li2026maxinfo} 
require encoding all candidate frames through a separate, often external vision encoder~\cite{zhai2023sigmoid,li2022blip,yu2023self, radford2021learning}, increasing inference cost to approximately $1.8\times$.

Notice that the advantage of {\ourmethod} is most pronounced in the long-video regime: on LVB videos longer than $20$ minutes, we outperform the strongest baseline (MaxInfo at $50.8$) by \textcolor{ForestGreen}{$\mathbf{+3.5}$} points ($54.3$); on MLVU videos longer than $30$ minutes, we outperform the strongest baselines (Iframe and Optical Flow, both at $52.1$) by \textcolor{ForestGreen}{$\mathbf{+2.1}$} points ($54.2$). 

\noindent\textbf{Combining with AKS.} 
{\ourmethod} is a plug-and-play framework that can be combined with other query-aware (\cmark) frame selection methods such as  AKS~\cite{tang2025adaptive}, that score candidate frames against the posed question. 
In this context, Swift Sampling serves as a high-speed pre-filter, narrowing the candidate pool before more computationally expensive query-based scoring takes place. As shown in Table~\ref{tab:main_results}, pre-filtering from 128 to 96 candidates (\textit{AKS+\ourmethod}) reduces the selection cost of AKS from $1.53\times$ to $1.43\times$. 
Crucially, this efficiency does not come at the expense of performance; rather, it improves overall accuracy across all three benchmarks, including gains of $+0.8$ points on MLVU and $+0.6$ points on LVB. 
Additionally, by initiating the process with a larger pool of $256$ frames and pre-filtering to $128$, we outperform the standard AKS ($256 \to 32$) by $+2.1\%$ on Video-MME and $+0.5\%$ on MLVU. Remarkably, this performance boost is achieved while simultaneously reducing the inference costs from $2.06\times$ to $1.59\times$. In summary, {\ourmethod} is a robust, architecture-agnostic technique that selects pivotal frames for long-form video processing.

In the following sections, we show the utility of {\ourmethod} in diverse video understanding tasks.
\begin{figure*}[t]
\centering
\includegraphics[width=\textwidth]{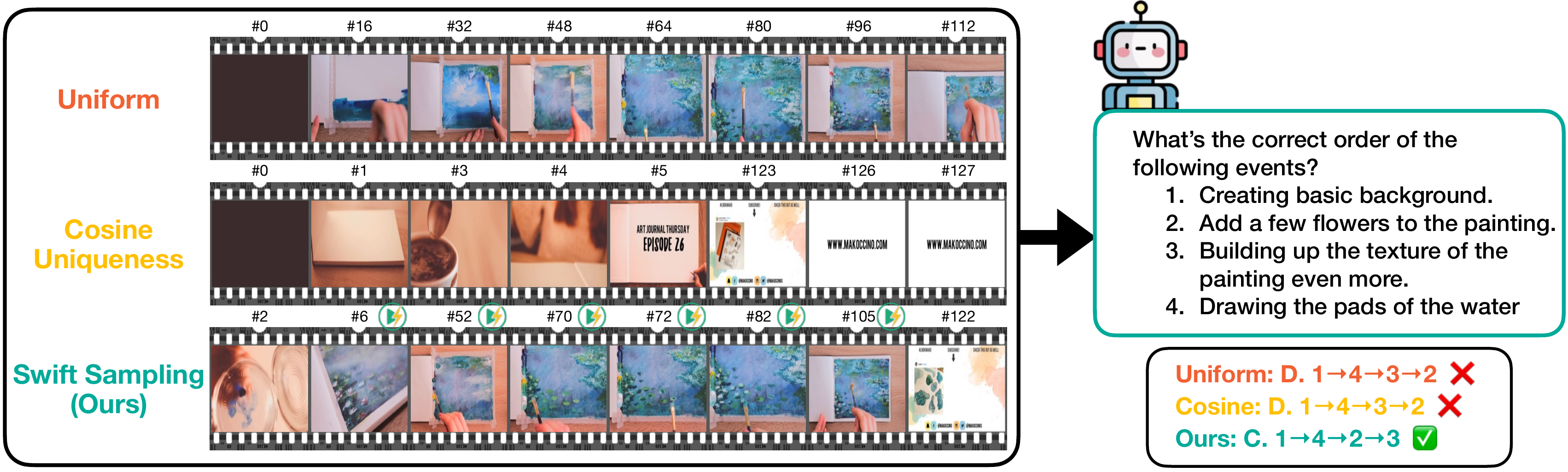}
\caption{\footnotesize \textbf{Qualitative comparison of frame selection} on a sample video from the Video-MME dataset, given a budget to select $8$ frames out of $128$. Answering the question requires identifying the temporal order of several visually similar but semantically distinct painting events: establishing the background, drawing the water-lily pads, adding flowers, and increasing texture. \textcolor{RedOrange}{Uniform sampling} captures the background and water-lily pads \textcolor{red}{but misses the frames showing the addition of flowers and texture}, leading to an incorrect answer. \textcolor{YellowOrange}{Cosine Uniqueness}~\cite{yuan2025unicomp} over-selects \textcolor{red}{visually salient but task-irrelevant frames}, such as title cards and end screens, which is especially harmful under a limited frame budget. \textcolor{Turquoise}{Swift Sampling} focuses on temporally informative changes in the painting progression, \textcolor{ForestGreen}{captures key intermediate stages} and enables correct temporal reasoning.}
\label{fig:qual_drawing}
\end{figure*}

\subsection{Application: Token Compression}
\begin{wraptable}{r}{0.45\textwidth}
\vspace{-0.4in}
\centering
\footnotesize
\caption{\footnotesize \textbf{{\ourmethod} for token compression} All methods produce $32$ frames as input to the video LLM. UniComp uses $32$ uniformly sampled frames. Replacing uniform sampling with Taylor selection (shown as \textit{+\ourmethod}) ($128 \to 32$ candidates) consistently improves the overall accuracy of UniComp~\cite{yuan2025unicomp} across all retain ratios $r$. \textbf{Bold}: best within block.}\label{tab:token_compression}\vspace{-0.1in}
\resizebox{0.45\textwidth}{!}{%
\begin{tabular}{lccccc}
\toprule
Method & \# Tokens & All & $\geq\!10$m & $\geq\!15$m & $\geq\!30$m \\
\midrule
Vanilla $32$f & $6272$ & 64.7 & 61.4 & 54.4 & 50.0 \\
\midrule
UniComp ($r=0.25$) & \multirow{2}{*}{$1568$} & 64.2 & 60.0 & 53.7 & 56.2 \\
\quad +\textit{{\ourmethod} (Ours)} & & \textbf{65.1} & \textbf{60.6} & \textbf{55.7} & \textbf{58.3} \\
\midrule
UniComp ($r=0.20$) & \multirow{2}{*}{$1254$} & 64.5 & \textbf{61.0} & 53.0 & 56.2 \\
\quad +\textit{{\ourmethod} (Ours)} & & \textbf{65.3} & 60.4 & \textbf{55.0} & \textbf{58.3} \\
\midrule
UniComp ($r=0.15$) & \multirow{2}{*}{$941$} & 64.4 & 59.8 & 57.0 & \textbf{62.5} \\
\quad +\textit{{\ourmethod} (Ours)} & & \textbf{66.0} & \textbf{61.8} & \textbf{57.7} & 56.2 \\
\midrule
UniComp ($r=0.10$) & \multirow{2}{*}{$627$} & 62.1 & 59.6 & \textbf{57.0} & \textbf{64.6} \\
\quad +\textit{{\ourmethod} (Ours)} & & \textbf{64.5} & \textbf{60.8} & 56.4 & 62.5 \\
\bottomrule
\end{tabular}
}\vspace{-0.2in}
\end{wraptable}
Token compression seeks to reduce visual token counts while preserving core video content. We integrate \ourmethod\ into UniComp~\cite{yuan2025unicomp}, the current state-of-the-art in this domain, by replacing its default uniform frame selection (indicated as \textit{+\ourmethod\ (Ours)}, $128 \to 32$ in Table~\ref{tab:token_compression}). Our intuition is that providing UniComp with more informative initial frames will yield superior results within the same frame budget. As shown in Table~\ref{tab:token_compression}, {\ourmethod} consistently boosts UniComp’s accuracy across all retention ratios $r$, achieving a peak gain of $+2.4$ points on MLVU ($r=0.10$). These results demonstrate that {\ourmethod} offers a \textit{drop-in improvement} for token compression pipelines.

\begin{wraptable}{r}{0.4\columnwidth}
\normalcolor
\centering
\footnotesize
\caption{\scriptsize \textbf{Swift Sampling for Video Captioning} improves accuracy across most categories in TempCompass~\cite{liu2024tempcompass}.}
\label{tab:captioning}
\resizebox{0.35\columnwidth}{!}{%
\scriptsize
\begin{tabular}{lcc}
\toprule
\textbf{Category} & \textbf{Uniform} & \textbf{Swift Sampling} \\
\midrule
Action           & 40.29 & 41.26 {\tiny $\mathbf{\textcolor{ForestGreen}{(+0.97)}}$} \\
Attribute Change & 38.52 & 34.44 {\tiny $\mathbf{\textcolor{Bittersweet}{(-4.08)}}$} \\
Direction        & 34.76 & 36.43 {\tiny $\mathbf{\textcolor{ForestGreen}{(+1.66)}}$} \\
Order            & 36.99 & 38.78 {\tiny $\mathbf{\textcolor{ForestGreen}{(+1.79)}}$} \\
Speed            & 12.37 & 13.14 {\tiny $\mathbf{\textcolor{ForestGreen}{(+0.77)}}$} \\
\bottomrule
\end{tabular}%
}
\vspace{0.2in}
\end{wraptable}

\subsection{Application: Video Captioning}
Video captioning requires interpreting a sequence of selected frames and generating a coherent natural language description. We evaluate {\ourmethod} on LLaVA-OneVision~\cite{li2024llava} on the TempCompass benchmark~\cite{liu2024tempcompass} on the task of caption generation, where improved frame selection is expected to yield more informative captions. For evaluation, we follow the original protocol: we prompt GPT-4o (2024-11-20)~\cite{hurst2024gpt} with the generated caption and ask it to answer a corresponding multiple-choice question. A correct answer indicates a correct caption, and vice versa. As shown in Table~\ref{tab:captioning}, {\ourmethod} improves captioning performance across nearly all categories, but struggles on attribute change.

\begin{wraptable}{r}{0.4\columnwidth}
\normalcolor
\centering
\footnotesize
\caption{\footnotesize \textbf{Per-task accuracy on Video-MME}} 
\label{tab:llava-task}
\resizebox{0.35\columnwidth}{!}{%
\footnotesize
\begin{tabular}{lcc}
\toprule
\textbf{Task Category} & \textbf{Uniform} & \textbf{\ourmethod} \\
\midrule
Action Reasoning       & 53.7 & 57.5 {\tiny $\mathbf{\textcolor{ForestGreen}{(+3.9)}}$} \\
Action Recognition     & 55.0 & 57.2 {\tiny $\mathbf{\textcolor{ForestGreen}{(+2.2)}}$} \\
Attribute Perception   & 74.8 & 71.2 {\tiny $\mathbf{\textcolor{Bittersweet}{(-3.6)}}$} \\
Counting Problem       & 37.7 & 35.4 {\tiny $\mathbf{\textcolor{Bittersweet}{(-2.2)}}$} \\
Information Synopsis   & 73.4 & 74.3 {\tiny $\mathbf{\textcolor{ForestGreen}{(+0.9)}}$} \\
OCR Problems           & 61.9 & 60.4 {\tiny $\mathbf{\textcolor{Bittersweet}{(-1.4)}}$} \\
Object Reasoning       & 55.1 & 55.1 {\tiny $\mathbf{(+0.0)}$} \\
Object Recognition     & 65.8 & 66.1 {\tiny $\mathbf{\textcolor{ForestGreen}{(+0.3)}}$} \\
Spatial Perception     & 59.3 & 63.0 {\tiny $\mathbf{\textcolor{ForestGreen}{(+3.7)}}$} \\
Spatial Reasoning      & 78.6 & 83.9 {\tiny $\mathbf{\textcolor{ForestGreen}{(+5.4)}}$} \\
Temporal Perception    & 63.6 & 60.0 {\tiny $\mathbf{\textcolor{Bittersweet}{(-3.6)}}$} \\
Temporal Reasoning     & 40.7 & 43.5 {\tiny $\mathbf{\textcolor{ForestGreen}{(+2.8)}}$} \\
\bottomrule
\end{tabular}%
}
\vspace{0.2in}
\end{wraptable}

\subsection{Applications: Other Downstream tasks in Video-MME}
We analyze performance by task-category on Video-MME to isolate {\ourmethod}'s core strengths. As seen from Table~\ref{tab:llava-task},  {\ourmethod} excels in reasoning-intensive tasks: Spatial Reasoning (\textcolor{ForestGreen}{$\mathbf{+5.4\%}$}), Action Reasoning (\textcolor{ForestGreen}{$\mathbf{+3.9\%}$}), Temporal Reasoning (\textcolor{ForestGreen}{$\mathbf{+2.8\%}$}), and Action Recognition (\textcolor{ForestGreen}{$\mathbf{+2.2\%}$}). We believe selecting frames that most distinctively capture the motion benefits these high-level tasks. By contrast, performance regresses in tasks requiring global temporal continuity, such as Temporal Perception ($-3.6\%$) and Counting ($-2.2\%$) (qualitative examples in Appendix \ref{app:qual_task_analysis}). We conjecture that these categories may demand uniform temporal coverage because even low-surprise regions of the video may carry task-relevant information, a requirement less suited to selective, surprise-based sampling.

\section{Analysis of Swift Sampling}\label{sec:analysis}\label{sec:4_feature_pooling}
Below, we present a thorough analysis of our method on LLaVA-OneVision~\cite{li2024llava}.

\begin{wraptable}{r}{0.5\columnwidth}
\centering
\footnotesize
\caption{\footnotesize  \textbf{Effect of spatial aggregation.} Patch grid for region-level pooling before computing the Taylor residual on MLVU. Global mean pooling provides the best balance across benchmarks. \textbf{Bold}: best.}
\label{tab:effect_aggregation}
\resizebox{0.45\columnwidth}{!}{%
\begin{tabular}{lcccc}
\toprule
Patch grid & Regions & MLVU & Video-MME & LVB \\
\midrule
\rowcolor{JungleGreen!8} Global mean (Ours) & 1 & 65.6 & \textbf{59.0} & \textbf{57.9} \\
$2 \times 2$ & 4 & 64.7 & 57.7 & 56.6 \\
$4 \times 4$ & 16 & \textbf{65.8} & 58.3 & 57.1 \\
$7 \times 7$ & 49 & 65.0 & 58.7 & 57.4 \\
$14 \times 14$ (all tokens) & 196 & 64.1 & 58.3 & 57.8 \\
\bottomrule
\end{tabular}
}
\vspace{0.1in}
\end{wraptable}

\noindent \textbf{Choice of feature pooling.}
We study how the spatial granularity of feature aggregation affects frame selection quality. LLaVA-OneVision~\cite{li2024llava} uses SigLIP~\cite{zhai2023sigmoid} as its vision encoder, which produces a $14 \times 14$ token grid per frame. We first aggregate the per-frame token grid into an $S \times S$ patch grid. Taylor residuals are computed for each grid's mean feature, then averaged to produce a single frame-level score. We sweep $S \in \{1, 2, 4, 7, 14\}$, ranging from global mean pooling ($S=1$, a single region summarizing all $196$ tokens) to no aggregation ($S=14$, each region containing exactly one token). As shown in Table~\ref{tab:effect_aggregation}, global mean pooling ($S=1$) achieves the best overall performance. We hypothesize that finer grids (e.g., $S = 14$) dilute the temporal signal presumably because the local residuals could be dominated by texture noise and camera jitter. 
By contrast, the frame-level mean captures coherent scene-level transitions more relevant to frame selection. Given its superior robustness $S=1$ is our default aggregation strategy.
\newpage
\begin{wraptable}{r}{0.4\columnwidth}  
\centering
\vspace{-0.1in}
\caption{\footnotesize \textbf{Effect of Feature Layer} $\ell=0$ yields the optimal trade-off between VQA accuracy and computational cost.}\vspace{-0.1in}
\label{tab:effect_layer}
\resizebox{0.35\columnwidth}{!}{%
\begin{tabular}{lcccc}
\toprule
Key layer & MLVU & Video-MME & LVB & FLOPs \\
\midrule
\rowcolor{JungleGreen!8} $\ell=0$  & 65.6 & 59.0 & \textbf{57.9} & \hspace{6pt} $1.02\times$ \twemoji{rocket}\\
$\ell=1$ & 64.3 & \textbf{59.1} & 56.3 & $1.05\times$ \\
$\ell=2$ & \textbf{65.8} & 58.4 & 57.1 & $1.07\times$ \\
$\ell=3$ & 65.5 & \textbf{59.1} & 57.7 & $1.09\times$ \\
\bottomrule
\end{tabular}
}
\end{wraptable}
\noindent \textbf{Choice of Feature Layer and Type.}\label{sec:feature_layer}
We study which layers and feature types yield the most predictable temporal dynamics, and thus, the most informative Taylor residuals for frame selection. 
We compare \emph{key} features from the self-attention projection $W_k$ against \emph{hidden} features (encoder block outputs).  As shown in Table~\ref{tab:effect_layer}, mean-pooled key features in the earliest layers of the vision encoder produce the lowest average residuals. We hypothesize that this is because early-layer features provide stable, low-level scene representations that evolve smoothly, making them highly predictable yet sensitive to sudden `surprises.' Deeper layers, though offer marginal gains, often prioritize holistic semantics over the motion dynamics required for precise temporal prediction. Thus, we use $\ell = 0$ as our default choice. We present a detailed visualization of all layers in Appendix. %

\noindent \textbf{Effect of Taylor Expansion Order.}
\begin{table*}[t]
\centering
\caption{\footnotesize \textbf{Analyses on Taylor expansion order and frame budget.}}
\label{tab:order_budget}
\begin{subtable}[t]{0.3\linewidth}
\centering
\caption{\footnotesize \textbf{Effect of Taylor expansion order $N$} on VQA accuracy across Video-MME, LongVideoBench (LVB), and MLVU. $N=3$ provides the best overall balance across benchmarks. \cradd{$N=0$ uses the previous feature as the prediction.}}
\label{tab:effect_order}
\resizebox{\linewidth}{!}{%
\begin{tabular}{lccc}
\toprule
Order & Video-MME & LVB & MLVU \\
\midrule
\cradd{$N=0$} & \cradd{57.2} & \cradd{56.3} & \cradd{64.2} \\
$N=1$ & 58.1 & 56.7 & 64.6 \\
$N=2$ & 58.7 & 56.6 & 64.9 \\
\rowcolor{JungleGreen!8} $N=3$ & \textbf{59.0} & \textbf{57.9} & 65.6 \\
$N=4$ & 58.0 & 57.2 & 65.7 \\
$N=6$ & 58.1 & 57.1 & \textbf{66.0} \\
$N=8$ & 58.3 & 57.4 & 65.9 \\
\bottomrule
\end{tabular}
}
\end{subtable}
\hfill
\begin{subtable}[t]{0.67\linewidth}
\centering
\caption{\footnotesize \textbf{Impact of frame budget ($K$) on MLVU VQA accuracy.} We compare uniform sampling and Cosine Uniqueness~\cite{yuan2025unicomp} against {\ourmethod}. Results are reported for the full MLVU dataset and specific long-video subsets ($\geq\!10$m, $\geq\!15$m, $\geq\!30$m). Gains over uniform sampling are indicated in the parentheses.}
\vspace{-0.1in}
\label{tab:effect_budget}
\resizebox{0.8\linewidth}{!}{%
\begin{tabular}{clcccc}
\toprule
$K$ & Frame Selection Method & \nogain{All} & \nogain{$\geq\!10\mathrm{m}$} & \nogain{$\geq\!15\mathrm{m}$} & \nogain{$\geq\!30\mathrm{m}$} \\
\midrule
\multirow{3}{*}{$32$} & Uniform                  & \nogain{64.7}            & \nogain{61.4}                     & \nogain{54.4}                     & \nogain{50.0} \\
 & Cosine Uniqueness                            & 65.4 {\tiny ($+0.7$)}     & \textbf{63.6}  {\tiny ($+2.2$)}   & \textbf{61.1} {\tiny ($+6.7$)} & 47.9  {\tiny ($-2.1$)} \\
\rowcolor{JungleGreen!8} &  {\ourmethod} (Ours) & \textbf{65.6} {\tiny ($+0.9$)} & 62.2 {\tiny ($+0.8$)}        & 58.2 {\tiny ($+3.8$)}         & \textbf{54.2} {\tiny ($+4.2$)} \\
\midrule
\multirow{3}{*}{$16$} & Uniform & \nogain{61.6} & \nogain{58.9} & \nogain{52.3} & \nogain{47.9} \\
 & Cosine Uniqueness & 61.4 {\tiny ($-0.2$)} & 57.9 {\tiny ($-1.0$)} & \textbf{56.4} {\tiny ($+4.1$)} & 47.9 {\tiny ($+0.0$)} \\
\rowcolor{JungleGreen!8} &  {\ourmethod} (Ours) & \textbf{63.9} {\tiny ($+2.3$)} & \textbf{60.0} {\tiny ($+1.1$)} & 53.0 {\tiny ($+0.7$)} & \textbf{50.0} {\tiny ($+2.1$)} \\
\midrule
\multirow{3}{*}{$8$} & Uniform & \nogain{58.6} & \nogain{57.9} & \nogain{49.0} & \nogain{50.0} \\
 & Cosine Uniqueness & 58.7 {\tiny ($+0.1$)} & 56.9 {\tiny ($-1.0$)} & \textbf{55.7} {\tiny ($+6.7$)} & \textbf{54.2} {\tiny ($+4.2$)} \\
\rowcolor{JungleGreen!8}  & {\ourmethod} (Ours) & \textbf{60.3} {\tiny ($+1.7$)} & \textbf{59.1} {\tiny ($+1.2$)} & 53.0 {\tiny ($+4.0$)} & \textbf{54.2} {\tiny ($+4.2$)} \\
\midrule
\multirow{3}{*}{$4$} & Uniform & \nogain{54.4} & \nogain{53.3} & \nogain{51.7} & \nogain{45.8} \\
 & Cosine Uniqueness & 55.5 {\tiny ($+1.1$)} & 54.9 {\tiny ($+1.6$)} & 49.7 {\tiny ($-2.0$)} & 54.2 {\tiny ($+8.4$)} \\
\rowcolor{JungleGreen!8}  & {\ourmethod} (Ours) & \textbf{56.7} {\tiny ($+2.3$)} & \textbf{55.3} {\tiny ($+2.0$)} & \textbf{55.0} {\tiny ($+3.3$)} & \textbf{58.3} {\tiny $\mathbf{\textcolor{ForestGreen}{(+12.5)}}$} \\
\midrule
\multirow{3}{*}{$2$} & Uniform & \nogain{51.8} & \nogain{50.8} & \nogain{49.0} & \nogain{43.8} \\
 & Cosine Uniqueness & 52.5 {\tiny ($+0.7$)} & 53.7 {\tiny ($+2.9$)} & \textbf{53.0} {\tiny ($+4.0$)} & \textbf{54.2} {\tiny ($+10.4$)} \\
\rowcolor{JungleGreen!8} &  {\ourmethod} (Ours) & \textbf{54.0} {\tiny ($+2.2$)} & \textbf{53.9} {\tiny ($+3.1$)} & 51.7 {\tiny ($+2.7$)} & \textbf{54.2} {\tiny $\mathbf{\textcolor{ForestGreen}{(+10.4)}}$} \\
\bottomrule
\end{tabular}
}\vspace{-0.1in}
\end{subtable}
\end{table*}
We investigate how the Taylor expansion order $N$ (as defined in Section~\ref{sec:4_setup}) affects frame selection quality. 
As shown in Table~\ref{tab:effect_order}, VQA accuracy improves sharply from $N=1$ to $N=3$ across all three benchmarks, after which performance saturates. While Video-MME and LVB show no significant gains, MLVU exhibits marginal improvements at $N=6$. This suggests that low-order terms effectively capture the majority of predictable local dynamics and higher-order derivatives provide diminishing returns for identifying temporal surprises. Thus, we adopt $N=3$ as the default, striking a balance between efficiency and predictive accuracy. 

\noindent \textbf{Effect of Frame Budget.} We evaluate the impact of the frame budget $K$ on VQA performance. From  Table~\ref{tab:effect_budget}, {\ourmethod} consistently outperforms uniform sampling across all frame budgets, with significant gains on longer videos and under highly constrained budgets. For videos exceeding 30 minutes, {\ourmethod} improves over uniform sampling by $\mathbf{\textcolor{ForestGreen}{+12.5}}$ points at $K=4$ and $\mathbf{\textcolor{ForestGreen}{+10.4}}$ points at $K=2$.  Thus, as the frame budget tightens, identifying temporal surprises becomes critical for model reasoning and {\ourmethod} offers a computationally efficient solution for it.

\section{Conclusion and Future Work}\label{sec:5}
We presented \textbf{Swift Sampling}, a training-free framework for long-video understanding that identifies keyframes via Taylor residuals. By reusing only the early layers of a VLM’s vision encoder, our method adds a negligible $0.02\times$ overhead -- $30\times$ less overhead than prior baselines. {\ourmethod}'s lightweight design and consistent top-performance over prior training-free baselines
makes it a seamless drop-in for pipelines like token compression and captioning.

Currently, {\ourmethod} is query-agnostic to prioritize efficiency. This may occasionally lead it to select visually surprising but semantically irrelevant frames, such as shot transitions. \cradd{Swift Sampling can also underperform when tasks require tracking repeated events or extracting information from visually static segments.} Future work will explore adapting the Taylor signal to be task-sensitive. Extending the framework to audio and spatio-temporal modalities to achieve a more holistic, context-aware understanding of video is another fruitful future direction.

\noindent\textbf{Acknowledgments.} We thank Seongheon Park and our fellow research group members at BU for helpful discussions and paper feedback.

\bibliographystyle{unsrt}
\bibliography{main}

\newpage
\appendix

\begin{center}
    \LARGE \textbf{Appendix}
    \vspace{1em}
\end{center}

\tableofcontents
\addtocontents{toc}{\protect\setcounter{tocdepth}{2}}

\section{Baseline details}
\label{sec:baseline_details}
This section provides implementation details for the baselines used in Sec.~\ref{sec:4}.

\subsection{Query-agnostic methods}
\paragraph{Cosine Uniqueness~\cite{yuan2025unicomp}.} The cosine uniqueness score is introduced in UniComp~\cite{yuan2025unicomp} to quantify information uniqueness and thereby capture intrinsic redundancy across frames. For each of the $T$ candidate frames, we extract a visual feature $f_t \in \mathbb{R}^d$ using a pretrained vision encoder~\cite{zhai2023sigmoid}, and define the uniqueness of frame $t$ as its mean cosine distance \textbf{to all remaining candidates}:
\begin{equation}
u_t = \frac{1}{T-1} \sum_{j \neq t} \bigl(1 - \cos(f_t, f_j)\bigr).
\end{equation}
A higher $u_t$ indicates a less redundant frame, so we select the top-$K$ frames ranked by $u_t$.

\paragraph{Frame Difference.}
This baseline scores each candidate frame by its pixel-level difference from the previous frame. For each frame $t \geq 1$, we compute the mean $\ell_1$ difference
\begin{equation}
d_t = \frac{1}{HWC} \lVert F_t - F_{t-1} \rVert_1,
\end{equation}
where $F_t \in \mathbb{R}^{H \times W \times C}$ denotes the raw RGB frame 
at time $t$. The top-$K$ frames ranked by $d_t$ are then selected.

\paragraph{I-Frame.}
This baseline uses codec-level statistics, taking the encoded packet size of each I-frame as a proxy for scene complexity. This is motivated by the tendency of video encoders to place I-frames near scene boundaries and allocate more bits to visually complex content. We use PyAV to parse the video bitstream and collect the temporal positions and packet sizes of all I-frames.If $n_I \geq K$, we select the top-$K$ I-frames ranked by packet size; otherwise, we retain all I-frames and fill the remaining $K-n_I$ slots with the largest P-frames by packet size. We sort the selected indices in temporal order before passing them to the video LLM.

\paragraph{P-Frame.}
Like the I-Frame baseline, this method uses codec-level statistics, taking the encoded packet size of each P- or B-frame as a proxy for motion and scene change because it reflects the motion-compensation residual relative to the reference frame. We use PyAV to parse the video bitstream and extract the frame type and packet size of each frame.
We then select the top-$K$ candidate frames ranked by packet size.

\paragraph{RAFT~\cite{teed2020raft}.}
This baseline scores each candidate frame by the magnitude of its optical flow relative to the previous frame, estimated with RAFT-small. 
We apply RAFT to consecutive frames within the candidate pool so that all baselines, including {\ourmethod}, operate on the same pool of $T$ frames.
We resize frames to $384 \times 384$ before feeding them into RAFT-small, matching the native input resolution of LLaVA-OneVision and LLaVA-Video. 
We pass consecutive pairs $(F_{t-1}, F_t)$ through RAFT-small and use the final refinement iteration as the estimated flow field. 
For frame $t \geq 1$, we define the motion score as the mean flow magnitude:
\begin{equation}
m_t = \frac{1}{HW} \sum_{p \in \Omega} \lVert \mathrm{flow}(F_{t-1}, F_t)[p] \rVert_2,
\end{equation}
where $\Omega$ denotes the set of pixel locations and $H \times W$ is the RAFT input resolution. We set $m_0 = 0$ because the first frame has no predecessor. We select the $K$ frames with the largest $m_t$ and sort the selected indices in temporal order before passing them to the video LLM.

\paragraph{DySeg~\cite{shen2025fastvid}.}
DySeg is the dynamic temporal segmentation component of FastVid~\cite{shen2025fastvid}, which partitions a video into temporally ordered segments by placing boundaries at local minima of adjacent-frame cosine similarity. We adapt DySeg for frame selection as follows:
\begin{enumerate}[label=(\roman*), leftmargin=*, itemsep=2pt]
    \item For each candidate frame, we extract a visual feature 
    $f_t \in \mathbb{R}^d$ using the pretrained vision 
    encoder~\cite{zhai2023sigmoid}.
    \item We compute adjacent-frame cosine similarities $s_i = \cos(f_{i-1}, f_i)$, yielding the sequence $\mathbf{S} = \{s_i\}_{i=1}^{T-1}$.
    \item We place segment boundaries at $\mathbf{B} = \mathbf{B}_1 \cup \mathbf{B}_2$, where $\mathbf{B}_1$ contains the indices of the $(c-1)$ smallest values in $\mathbf{S}$ and $\mathbf{B}_2 = \{i \mid s_i < \tau\}$ contains all transitions whose similarity falls below $\tau$.
    \item We allocate the $K$-frame budget to the resulting segments in proportion to their lengths.
    \item Within each segment, we select the frames whose features are 
    closest to the segment centroid.
\end{enumerate}
We follow FastVid~\cite{shen2025fastvid} and set $c = 8$ and $\tau = 0.9$.

\subsection{Query-aware methods}
\paragraph{AKS~\cite{tang2025adaptive}.}
AKS is a query-aware sampling method that adaptively partitions a video according to query relevance scores over time. We use the official implementation with BLIP-ITM-Large~\cite{li2022blip} as the relevance scorer. For each candidate frame $t$, we feed the raw RGB frame $F_t$ and query $q$ into BLIP-ITM and use the matching probability as the relevance score:
\begin{equation}
s_t = \mathrm{softmax}\bigl(\mathrm{BLIP\text{-}ITM}(F_t, q)\bigr)_{\text{match}}.
\end{equation}
AKS then performs adaptive segment splitting on the score sequence $\{s_t\}$. Starting from the full timeline, AKS recursively divides a segment in half whenever it lacks a clear relevance peak. Segment peakedness is measured by $s_{\text{top}} - s_{\text{all}}$, where $s_{\text{top}}$ and $s_{\text{all}}$ denote the mean scores of the top-scoring frames and all frames within the segment, respectively. A segment is kept intact when $s_{\text{top}} - s_{\text{all}} \geq s_{\text{thr}}$; otherwise, recursion continues up to a maximum depth of $L$. Finally, AKS allocates the $K$-frame budget to the resulting segments in proportion to their lengths and selects the highest-scoring frames within each segment. We follow the original paper and set $s_{\text{thr}} = 0.1$ and $L = 5$.

\section{Swift Sampling: Implementation Details}\label{sec:impl_details}
All experiments are inference-only and do not require training. We run experiments on NVIDIA GPUs with at least 48GB of memory and use the lmms-eval library~\cite{zhang2024lmmsevalrealitycheckevaluation,lmms_eval2024} for benchmark evaluation.
We evaluate on the standard benchmarks: the multi-choice subset of the MLVU dev set~\cite{zhou2025mlvu} (\texttt{mlvu}, $2{,}174$ questions), the full Video-MME benchmark~\cite{fu2025video} ($2{,}700$ questions), and the validation split of LongVideoBench~\cite{wu2024longvideobench} (\texttt{longvideobench\_val\_v}, $1{,}337$ questions).
For early frames with insufficient history for the target Taylor order, we use the highest order supported by the available preceding frames. Specifically, the second frame uses zeroth-order prediction with one preceding frame, the third uses first-order prediction with two preceding frames, and the fourth uses second-order prediction with three preceding frames. From the fifth frame onward, we apply the full $N{=}3$ expansion.

\section{Additional analysis}\label{app:additional_analysis}

\subsection{Choice of feature layer}\label{app:feature_layer}
\begin{figure}[t]
\centering
\includegraphics[width=\columnwidth]{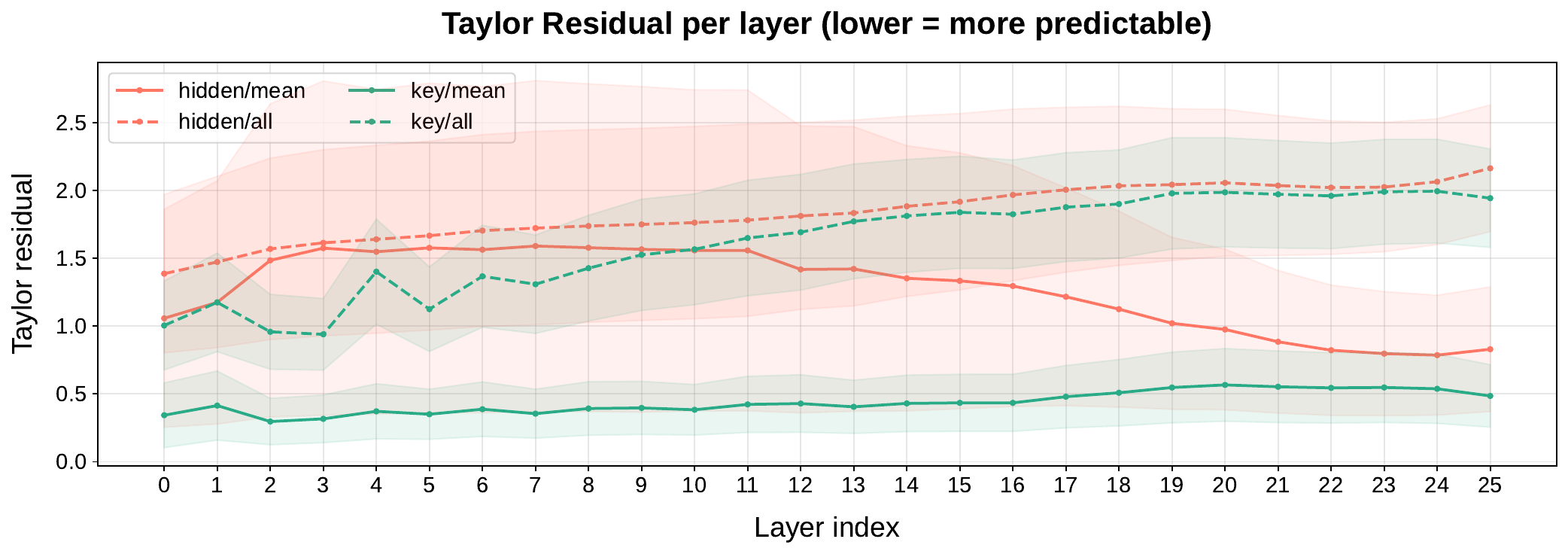}\vspace{-10pt}
\caption{\footnotesize \textbf{Taylor residual across different layers.} We compare \emph{key} features (green) and hidden output features (orange), with (solid) and without (dashed) spatial mean pooling. Key features with mean pooling yield the lowest residual, and the residual is smallest at the earliest layers, indicating that they are the most predictable from their temporal context. We use layer $\ell=0$ key features throughout our experiments. Shaded regions indicate standard deviation over 100 videos.} 
\vspace{-10pt}
\label{fig:effect_layer}
\end{figure} 
We provide a layer-by-layer analysis supporting the choice of $\ell{=}0$ in Sec.~\ref{sec:feature_layer}.
For each layer $\ell \in \{0,1,\ldots,25\}$ of the SigLIP~\cite{zhai2023sigmoid} vision encoder in LLaVA-OneVision~\cite{li2024llava}, we extract per-frame features under four configurations: (i) \emph{key/mean}: self-attention key projection $W_k$ with mean pooling ($S{=}1$); (ii) \emph{key/all}: $W_k$ with all tokens ($S{=}14$); (iii) \emph{hidden/mean}: post-MLP output with mean pooling; and (iv) \emph{hidden/all}: post-MLP output with all tokens.
We compute Taylor residuals as in Sec.~\ref{sec:4_feature_pooling}, average the per-frame scores within each video, and report the mean over $100$ randomly sampled videos.

Figure~\ref{fig:effect_layer} shows two clear trends.
First, key features yield lower residuals than hidden features at every layer, consistent with prior findings that attention keys carry stable structural information~\cite{amir2021deep,pan2024dissecting,yang2025visionzip}. 
Second, residuals are smallest in the earliest layers and grow with depth, regardless of feature type.
We hypothesize that latent features in the earliest layers provide a stable, low-level representation of the scene that evolves smoothly over time, making them highly predictable from their immediate temporal context.
Spatial mean pooling further reduces the residual at all layers, consistent with our finding in Section~\ref{sec:4_feature_pooling} that frame-level summaries suppress local noise. Based on these observations, we use $\ell{=}0$ key features with mean pooling as our default throughout all experiments, which also yields the lowest computational overhead.

\subsection{Effect of feature type }
\begin{table}[t]
\centering
\caption{\footnotesize \textbf{Effect of feature type from layer $\ell=0$ on VQA accuracy.} Key features achieve the best overall balance. \textbf{Bold}: best; \underline{underline}: second-best.}
\label{tab:effect_feature_type}
\resizebox{0.5\linewidth}{!}{%
\begin{tabular}{lccc}
\toprule
Feature & MLVU & Video-MME & LVB \\
\midrule
output (post-MLP hidden) & 65.2 & 58.7 & \underline{57.7} \\
query & 65.2 & 58.7 & \textbf{57.9} \\
value & \underline{65.3} & \underline{58.9} & \textbf{57.9} \\
\rowcolor{JungleGreen!8} key (Ours) & \textbf{65.6} & \textbf{59.0} & \textbf{57.9} \\
\bottomrule
\end{tabular}
}
\end{table}
We study which layer output provides the most useful residual by comparing key, query, value, and post-MLP hidden features at layer $\ell{=}0$.
Table~\ref{tab:effect_feature_type} shows that all four feature types yield similar accuracy, with key features achieving the best overall balance across benchmarks. 
This is consistent with Sec.~\ref{app:feature_layer}, where key features produce the lowest residuals across layers. We hypothesize that key features capture stable structural information used by attention~\cite{amir2021deep,pan2024dissecting,yang2025visionzip}, making their temporal dynamics smoother and more predictable. We therefore use key features as our default.

\begin{craddedblock}
\subsection{Controlled Diagnostics}
\label{app:cr_controlled}

We conducted two diagnostics using procedurally rendered 128-frame clips with controlled trajectories and events. The first examines whether higher-order Taylor terms explain smooth motion; the second evaluates whether the residual localizes abrupt events. Figure~\ref{fig:cr_trajectories} shows representative frames from the abrupt-event clips and their matched controls, and Figure~\ref{fig:cr_controlled} shows the corresponding Taylor residuals.

\paragraph{Higher-order Taylor terms explain smooth motion.}
For a fair comparison across Taylor orders, we report normalized residuals and use temporally shuffled clips as a control (Table~\ref{tab:cr_smooth}).

\begin{table}[htbp]
  \centering
  \ifcrreview\color{CRAdded}\captionsetup{font+={color=CRAdded},labelfont+={color=CRAdded}}\fi
  \small
  \caption{Normalized Taylor residuals on smooth trajectories and a temporally shuffled control. Lower residuals indicate better prediction.}
  \label{tab:cr_smooth}
  \begin{tabular}{lrrrr}
    \toprule
    Trajectory & $N=0$ & $N=1$ & $N=2$ & $N=3$ \\
    \midrule
    Constant velocity & 1.000 & 0.642 & 0.533 & \textbf{0.470} \\
    Constant acceleration & 1.000 & 0.673 & 0.574 & \textbf{0.511} \\
    Constant jerk & 1.000 & 0.699 & 0.604 & \textbf{0.533} \\
    Temporally shuffled control & 1.000 & 1.007 & 1.005 & 1.005 \\
    \bottomrule
  \end{tabular}
\end{table}

Higher-order prediction reduces the normalized residual by roughly half for smooth trajectories, whereas this reduction disappears after temporal shuffling. This shows that the higher-order Taylor terms capture temporally predictable feature evolution.

\paragraph{Abrupt deviations produce localized residual peaks.}
For each event condition, we generated 60 pairs of 128-frame clips, each consisting of one event clip and one matched no-event control. The paired clips share the same seed and are pixel-identical before the known event time $t^{*}$. We report Recall@8, defined as the percentage of clips in which at least one of the eight selected frames falls within the three-frame event window $\{t^{*}-1,t^{*},t^{*}+1\}$. For the matched-control clips, we evaluate the same nominal event window as in the paired event clip. Table~\ref{tab:cr_events} reports the results.

\begin{table}[htbp]
  \centering
  \ifcrreview\color{CRAdded}\captionsetup{font+={color=CRAdded},labelfont+={color=CRAdded}}\fi
  \small
  \setlength{\tabcolsep}{3pt}
  \caption{Recall@8 (\%) for controlled abrupt events. The Swift control column evaluates matched no-event clips; all other columns evaluate event clips. CU, FD, and OF denote Cosine Uniqueness, frame difference, and optical flow, respectively.}\vspace{0.1in}
  \label{tab:cr_events}
  \begin{tabular}{lrrrrrr}
    \toprule
    Event & \shortstack{Swift\\(event)} & \shortstack{Swift\\(control)} & CU & FD & OF & Uniform \\
    \midrule
    Object appearance & \textbf{100} & 0 & 32 & 100 & 100 & 22 \\
    Object disappearance & \textbf{100} & 0 & 32 & 100 & 100 & 22 \\
    Hard scene cut & \textbf{100} & 18 & 63 & 100 & 100 & 22 \\
    Direction change & \textbf{100} & 18 & 58 & 13 & 12 & 22 \\
    Acceleration to stop & \textbf{88} & 3 & 0 & 48 & 28 & 22 \\
    Camera-pan reversal & \textbf{100} & 22 & 60 & 23 & 15 & 22 \\
    Velocity to stop & \textbf{70} & 18 & 0 & 17 & 10 & 22 \\
    \midrule
    Mean & \textbf{94} & 11 & 35 & 57 & 52 & 22 \\
    \bottomrule
  \end{tabular}
\end{table}

The direction-change condition directly separates unpredictability from motion magnitude. The event and matched-control clips have nearly identical optical flow (0.197 vs.\ 0.199) and pixel difference (0.26 vs.\ 0.26), but Swift Sampling selects the event window in 100\% of event clips, compared with 18\% around the same nominal time in the matched controls. Frame difference and optical flow remain low at 13\% and 12\%, respectively. Thus, the Taylor residual responds to a violation of the preceding trajectory, not simply to the amount of visual change.

\begin{figure}[htbp]
  \centering
  \ifcrreview\color{CRAdded}\captionsetup{font+={color=CRAdded},labelfont+={color=CRAdded}}\fi
  \includegraphics[width=\linewidth]{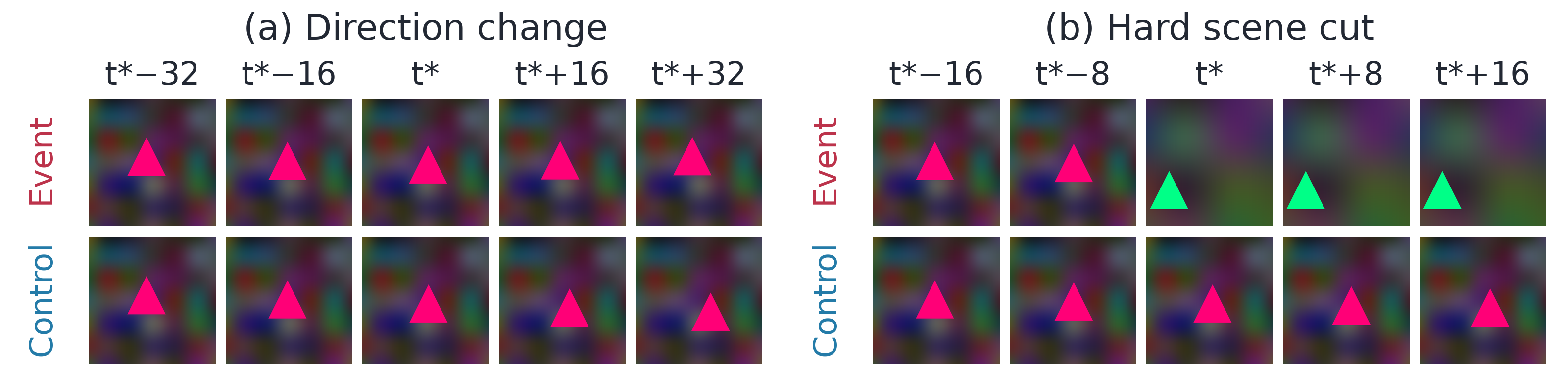}
  \caption{Representative frames from controlled clips with a direction change or a hard scene cut. For each example, the top row shows the event clip and the bottom row shows its matched no-event control. Frame times are shown relative to the known event time $t^*$.}
  \label{fig:cr_trajectories}
\end{figure}

\begin{figure}[htbp]
  \centering
  \ifcrreview\color{CRAdded}\captionsetup{font+={color=CRAdded},labelfont+={color=CRAdded}}\fi
  \includegraphics[width=0.7\linewidth]{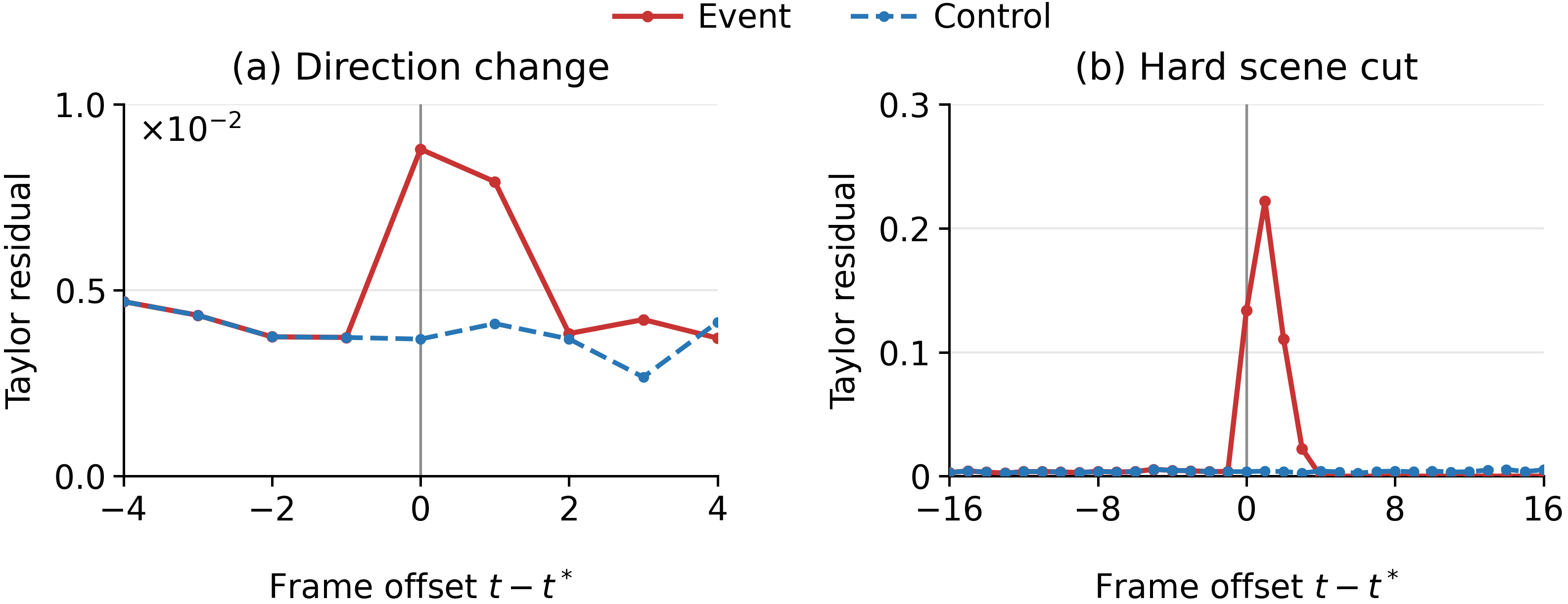}
  \caption{Taylor residuals ($N=3$) for the clips in Figure~\ref{fig:cr_trajectories}. Solid red and dashed blue curves correspond to event and matched no-event control clips, respectively. Vertical lines mark the known event time $t^*$.}
  \label{fig:cr_controlled}
\end{figure}

\paragraph{Controlled events on real-video segments.}
We further validated this behavior using MLVU and LongVideoBench segments as real-video backgrounds. Swift Sampling localized controlled scene cuts in 93\% of clips, compared with 18\% for matched controls, and controlled object insertions in 62\%, compared with 17\% (Table~\ref{tab:cr_real_background}).

\begin{table}[htbp]
  \centering
  \ifcrreview\color{CRAdded}\captionsetup{font+={color=CRAdded},labelfont+={color=CRAdded}}\fi
  \small
  \caption{Localization rates (\%) for controlled events inserted into real-video backgrounds from MLVU and LongVideoBench.}\vspace{0.1in}
  \label{tab:cr_real_background}
  \begin{tabular}{lrr}
    \toprule
    Controlled intervention & Event clips & Matched controls \\
    \midrule
    Scene cut & 93 & 18 \\
    Object insertion & 62 & 17 \\
    \bottomrule
  \end{tabular}
\end{table}
\end{craddedblock}

\begin{craddedblock}
\subsection{Temporal Stride Ablation}
\label{app:cr_temporal}

The temporal stride $s$ denotes the index spacing between the preceding candidate frames used for Taylor prediction. Increasing $s$ allows the prediction to use a broader temporal context. We tested temporal strides $s\in\{1,2,4,8\}$ on LongVideoBench, keeping $T=128\rightarrow K=32$, $N=3$, and all other settings fixed. The corresponding accuracies are 57.9, 57.7, 57.5, and 57.5, respectively (Table~\ref{tab:cr_stride}). This shows that Swift Sampling is robust to the temporal stride.

\begin{table}[htbp]
  \centering
  \ifcrreview\color{CRAdded}\captionsetup{font+={color=CRAdded},labelfont+={color=CRAdded}}\fi
  \small
  \caption{Temporal-stride ablation on LongVideoBench with $T=128\rightarrow K=32$ and $N=3$.}\vspace{0.1in}
  \label{tab:cr_stride}
  \begin{tabular}{lrrrr}
    \toprule
    Temporal stride $s$ & 1 & 2 & 4 & 8 \\
    \midrule
    Accuracy & \textbf{57.9} & 57.7 & 57.5 & 57.5 \\
    \bottomrule
  \end{tabular}
\end{table}

We examined the frames selected by Swift Sampling at temporal strides $s=1$ and $s=8$ across several videos (Figure~\ref{fig:cr_stride}). First, the two strides often selected the same informative frame, indicating that the residual signal used to identify informative frames is stable across temporal strides. Second, in the cases where the selections diverged, we manually verified that the frame chosen at $s=8$ was temporally adjacent to the one chosen at $s=1$ and depicted the same event. This suggests that informative events can span multiple neighboring frames, allowing nearby selections within the same temporal window to provide comparable evidence. We retain $s=1$ as the default since it preserves the finest-grained local temporal context and yields the best accuracy.

\begin{figure}[htbp]
  \centering
  \ifcrreview\color{CRAdded}\captionsetup{font+={color=CRAdded},labelfont+={color=CRAdded}}\fi
  \includegraphics[width=0.8\linewidth]{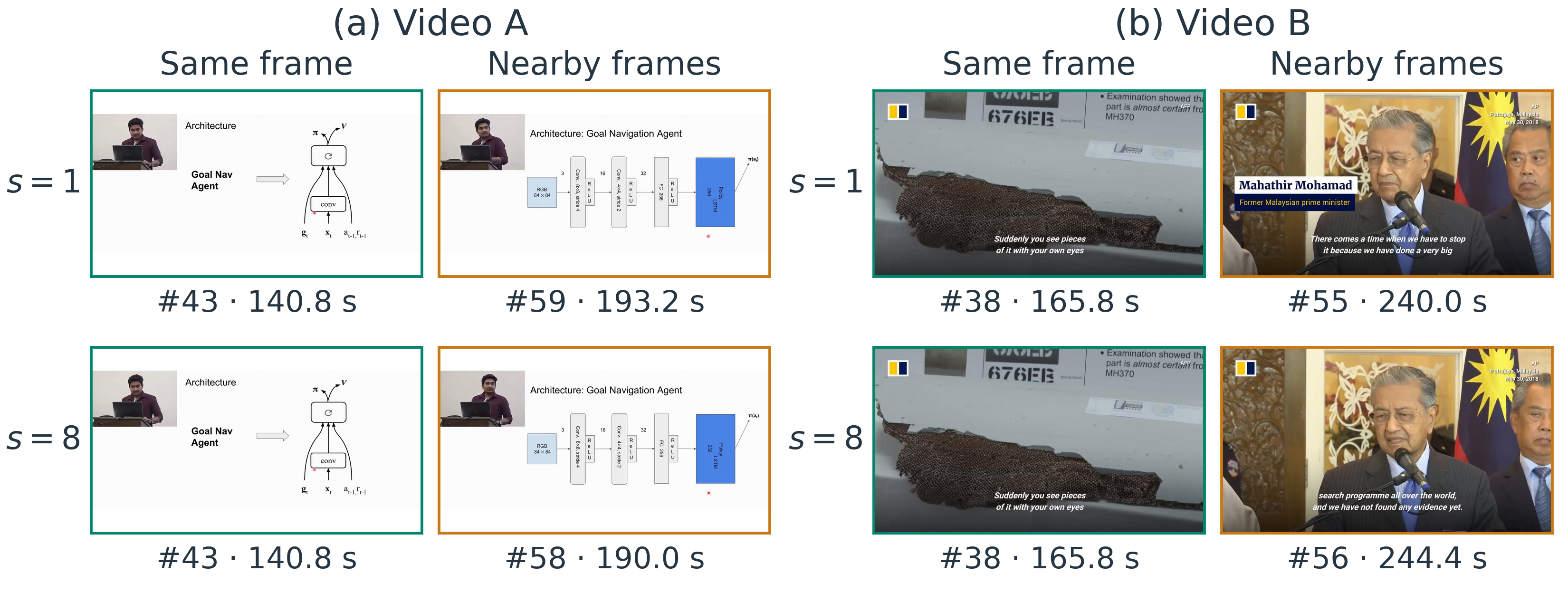}
  \caption{Qualitative examples of Swift Sampling's frame selections at temporal strides $s=1$ and $s=8$ for two videos, with $T=128\rightarrow K=32$ and $N=3$. Green-bordered pairs show the same candidate selected at both strides, while orange-bordered pairs show neighboring candidates depicting the same event. Numbers below the frames denote indices in the candidate sequence.}
  \label{fig:cr_stride}
\end{figure}
\end{craddedblock}

\begin{craddedblock}
\subsection{Temporal Smoothing}
We evaluated whether temporal smoothing improves frame selection by reducing short-term fluctuations in features or residuals. We tested three variants on LongVideoBench. Feature smoothing averages features over temporal windows before computing the Taylor residual. Residual smoothing smooths the residual sequence before frame selection. Step-wise feature averaging averages features within each temporal step used for Taylor prediction. We tested window sizes $w\in\{2,4\}$ while keeping all other settings fixed (Table~\ref{tab:cr_smoothing}).
\begin{table}[htbp]
  \centering
  \ifcrreview\color{CRAdded}\captionsetup{font+={color=CRAdded},labelfont+={color=CRAdded}}\fi
  \small
  \caption{Temporal-smoothing ablation on LongVideoBench.}\vspace{0.1in}
  \label{tab:cr_smoothing}
  \begin{tabular}{lrr}
    \toprule
    Method & Window $w$ & Accuracy \\
    \midrule
    No temporal smoothing & -- & \textbf{57.9} \\
    Feature smoothing & 2 & 57.5 \\
    Feature smoothing & 4 & 56.5 \\
    Residual smoothing & 2 & 57.5 \\
    Residual smoothing & 4 & 56.8 \\
    Step-wise feature averaging & 2 & 57.5 \\
    Step-wise feature averaging & 4 & 57.1 \\
    \bottomrule
  \end{tabular}
\end{table}

None of the smoothing variants improves performance. Mild smoothing with $w=2$ reduces accuracy by only 0.4 points, while $w=4$ degrades accuracy by up to 1.4 points. Temporal smoothing may attenuate informative event boundaries, so we do not apply it by default.
\end{craddedblock}

\subsection{Local Maxima vs.\ Top-$K$ Selection}
\begin{table}[t]
\centering
\caption{\footnotesize \textbf{Selection rule ablation.}
VQA accuracy on Video-MME, LongVideoBench (LVB), and MLVU under $T{=}128\!\to\!K{=}32$ with LLaVA-OneVision.
All methods use the same Taylor residuals ($N{=}3$); only the selection rule differs.
For window-based top-$K$, the video is divided into $K$ windows (stride $T/K$), and $\mathrm{wm}$ controls the degree of overlap by scaling the window size ($\mathrm{wm}{=}1$: no overlap).
\cradd{Radius suppression greedily selects the highest-residual remaining frame and excludes candidates within a temporal radius $\rho$ of each selected frame.}
Bold indicates the best result in each column.}\vspace{0.1in}
\label{tab:effect_topk}
\footnotesize
\begin{tabular}{llccc}
\toprule
Family & Variant & MLVU & Video-MME & LVB \\
\midrule
Baseline & Uniform sampling & 64.7 & 58.3 & 55.3 \\
\midrule
Global top-$K$ & Top-$K$ & 64.6 & 56.9 & 56.5 \\
\midrule
\multirow{4}{*}{Window-based top-$K$}
 & $\mathrm{wm}{=}1$ & 64.7 & 58.8 & 57.2 \\
 & $\mathrm{wm}{=}2$ & 64.6 & 57.1 & 55.7 \\
 & $\mathrm{wm}{=}3$ & 65.3 & 57.1 & 55.4 \\
 & $\mathrm{wm}{=}4$ & 64.8 & 57.3 & 55.8 \\
\midrule
\cradd{Radius suppression}
 & \cradd{$\rho=1$} & \cradd{--} & \cradd{--} & \cradd{57.1} \\
 & \cradd{$\rho=2$} & \cradd{--} & \cradd{--} & \cradd{56.2} \\
 & \cradd{$\rho=3$} & \cradd{--} & \cradd{--} & \cradd{55.9} \\
\midrule
\rowcolor{JungleGreen!8} Local maxima & \ourmethod\ (Ours) & \textbf{65.6} & \textbf{59.0} & \textbf{57.9} \\
\bottomrule
\end{tabular}
\end{table}

In Sec.~\ref{sec:3.2}, we select local maxima of $\{r_t\}$ because the absolute scale of $r_t$ depends on local trajectory dynamics. We compare this choice with two alternatives: (i) global top-$K$, which selects the $K$ largest values of $r_t$ over the entire video, and (ii) window-based top-$K$, which selects the highest-residual frame within temporal windows of stride $T/K$, with overlap controlled by a window multiplier $\mathrm{wm}\in\{1,2,3,4\}$.
Table~\ref{tab:effect_topk} shows that local maxima consistently outperform both global and window-based top-$K$ across all benchmarks. Window-based top-$K$ is also highly sensitive to the choice of $\mathrm{wm}$, with accuracy varying by up to $1.7$ points on Video-MME and LVB. Local maxima avoid this hyperparameter while delivering the best overall accuracy. We therefore adopt local maxima as the default.

\begin{craddedblock}
\paragraph{Local Maxima vs.\ Hard Temporal Separation.}
\label{app:cr_selection}

Local-maxima selection is designed to identify peaks relative to their local temporal context, rather than to guarantee a fixed minimum distance between selected frames. We additionally evaluate radius-based suppression on LongVideoBench under the same $T=128\rightarrow K=32$ and $N=3$ setting. After selecting the highest-residual frame, we exclude all candidates within radius $\rho$ and repeat until $K=32$ frames are selected. Table~\ref{tab:effect_topk} compares $\rho\in\{1,2,3\}$ with the other selection strategies.

Local-maxima selection achieves the best accuracy. Even the mildest suppression, $\rho=1$, reduces accuracy by 0.8 points, with larger radii causing further degradation. These results indicate that a fixed exclusion radius can remove nearby but distinct surprises, whereas local-maxima selection adapts to the temporal density of residual peaks.

\end{craddedblock}

\subsection{Filling Strategies for Local Maxima Selection}
\begin{table}[t]
\centering
\caption{\footnotesize \textbf{Filling strategy ablation.}
VQA accuracy on Video-MME, LongVideoBench (LVB), and MLVU under $N{=}128\!\to\!K{=}32$ with LLaVA-OneVision.
All methods use the same Taylor residuals and local-maxima selection; only the strategy for filling the remaining frames differs.
Residual-based filling selects additional frames by top residual, while uniform filling distributes them evenly over time.
Bold indicates the best result in each column.}
\label{tab:effect_fill}
\footnotesize
\begin{tabular}{lccc}
\toprule
Strategy & MLVU & Video-MME & LVB \\
\midrule
Uniform filling & 65.2 & 58.7 & 57.6 \\
\rowcolor{JungleGreen!8} Residual-based filling (Ours) & \textbf{65.6} & \textbf{59.0} & \textbf{57.9} \\
\bottomrule
\end{tabular}
\end{table}

In Sec.~\ref{sec:3.2}, we select frames based on local maxima of the Taylor residuals. 
When the number of detected local maxima $M$ is smaller than the target budget $K$, we fill the remaining $K-M$ slots with the unselected frames that have the largest Taylor residuals. Our default choice is to select the remaining frames by top-$K$ residual, as frames with high residuals are more likely to be informative even if they are not local maxima.
To validate this design, we compare our method against a Uniform filling strategy, which distributes the remaining slots evenly across temporal gaps between the detected local maxima. As shown in Table~\ref{tab:effect_fill}, our method consistently outperforms Uniform filling across all benchmarks. 
We therefore adopt this strategy as the default.

\subsection{Additional baselines}\label{app:additional_baselines}

In Table~\ref{tab:main_results} we compare {\ourmethod} against existing query-agnostic frame selection baselines using their default configurations (top-$K$ selection, last-layer ViT features). A natural question is whether the design choices that benefit {\ourmethod}, local-maxima selection and early-layer features, would also benefit these baselines. 
We address this in two controlled studies on LLaVA-OneVision, isolating each design choice independently.

\paragraph{Selection strategy: Top-$K$ vs.\ Local Maxima.}
\begin{table}[t]
\centering
\caption{\footnotesize \textbf{Top-$K$ vs.\ Local Maxima selection on Optical Flow.} 
We apply both selection strategies to optical flow~\cite{teed2020raft} on LLaVA-OneVision. 
Optical flow shows only marginal changes between selection rules, in contrast to the 
substantial gain that local-maxima selection provides for {\ourmethod} 
(see Table~\ref{tab:effect_topk}). 
This indicates that the benefit of local-maxima selection depends on whether 
the underlying score conditions on temporal context.}
\label{tab:baseline_selection}
\resizebox{0.6\columnwidth}{!}{%
\begin{tabular}{l c c c c c}
\toprule
Method & Selection & \# Frames & Video-MME & LVB & MLVU \\
\midrule
Optical Flow~\cite{teed2020raft}  & Top-$K$       & 128$\to$32 & 56.5 & 56.9 & 62.9 \\
Optical Flow~\cite{teed2020raft}  & Local Maxima  & 128$\to$32 & 56.6 & 57.2 & 63.6 \\
\midrule
\rowcolor{JungleGreen!8} {\ourmethod} (Ours) & Local Maxima & 128$\to$32 & \textbf{59.0} & \textbf{57.9} & \textbf{65.6} \\
\bottomrule
\end{tabular}
}
\vspace{-10pt}
\end{table}
{\ourmethod} adopts local-maxima selection since $r_t$ is computed from preceding frames, and thus its absolute magnitude depends on the local dynamics of the trajectory (Sec.~\ref{sec:3.2}).
In contrast, baselines such as Cosine Uniqueness~\cite{yuan2025unicomp} and optical flow~\cite{teed2020raft} produce scores that do not condition on temporal context, thus it is natural to use global top-$K$~\cite{yuan2025unicomp}.
For completeness, we additionally apply local-maxima selection to optical flow following the same procedure as in Sec.~\ref{sec:3.2}. As shown in Table~\ref{tab:baseline_selection}, local-maxima selection brings only marginal improvements to optical flow ($+0.1$ on Video-MME, $+0.3$ on LVB, $+0.7$ on MLVU), whereas {\ourmethod} obtains substantially larger gains under the same selection rule ($+2.1$ on Video-MME, $+1.4$ on LVB, $+1.0$ on MLVU; Table~\ref{tab:effect_topk}). 

\paragraph{Feature layer selection.}
\begin{table}[t]
\centering
\caption{\footnotesize \textbf{Effect of feature layer.} 
We compare two layers of the SigLIP encoder for Cosine Uniqueness~\cite{yuan2025unicomp} on LLaVA-OneVision: the last layer ($\ell{=}25$), which is the default used in~\cite{yuan2025unicomp}, and the early layer ($\ell{=}0$) adopted by {\ourmethod}.
Cosine Uniqueness consistently performs better with the last layer, indicating that the use of early-layer features is specific to {\ourmethod} and does not transfer to other selection methods.
\textbf{Bold}: best.}
\label{tab:baseline_layer}
\resizebox{\columnwidth}{!}{%
\begin{tabular}{l c c|
  >{\columncolor{RedOrange!4}}c >{\columncolor{RedOrange!8}}c >{\columncolor{RedOrange!14}}c >{\columncolor{RedOrange!22}}c |
  >{\columncolor{Goldenrod!4}}c >{\columncolor{Goldenrod!8}}c >{\columncolor{Goldenrod!14}}c >{\columncolor{Goldenrod!22}}c |
  >{\columncolor{JungleGreen!4}}c >{\columncolor{JungleGreen!8}}c >{\columncolor{JungleGreen!14}}c >{\columncolor{JungleGreen!22}}c}
\toprule
\multirow{2}{*}{Method} & \multirow{2}{*}{Layer} & \multirow{2}{*}{\# Frames} & \multicolumn{4}{c}{Video-MME} & \multicolumn{4}{c}{LVB} & \multicolumn{4}{c}{MLVU} \\
\cmidrule(lr){4-7} \cmidrule(lr){8-11} \cmidrule(lr){12-15}
& & & Short & Medium & Long & Overall & $\geq$10m & $\geq$20m & $\geq$30m & Overall & $\geq$10m & $\geq$15m & $\geq$30m & Overall \\
\midrule
Cosine Uniqueness~\cite{yuan2025unicomp} & $\ell{=}25$  & 128$\to$32 & 65.3 & 54.7 & 47.0 & 55.7 & 47.0 & 47.1 & 46.3 & 52.5 & \textbf{63.6} & \textbf{61.1} & 47.9 & 65.4 \\
Cosine Uniqueness~\cite{yuan2025unicomp} & $\ell{=}0$   & 128$\to$32 & 62.9 & 52.6 & 48.6 & 54.7 & \textbf{54.7} & 47.0 & 47.1 & 46.3 & 59.1 & 53.7 & 47.9 & 62.7 \\
\midrule
{\ourmethod} (Ours)                      &$\ell{=}0$   & 128$\to$32 & \textbf{71.0} & \textbf{56.9} & \textbf{49.2} & \textbf{59.0} & 51.6 & \textbf{54.3} & \textbf{50.9} & \textbf{57.9} & 62.2 & 58.2 & \textbf{54.2} & \textbf{65.6} \\
\bottomrule
\end{tabular}
}
\end{table}
{\ourmethod} uses early-layer ($\ell{=}0$) SigLIP features, as early layers of SigLIP yield smoother temporal trajectories and thus more predictable residuals (Sec.~\ref{app:feature_layer}).
We test whether this choice transfers to Cosine Uniqueness~\cite{yuan2025unicomp}, comparing its default last-layer features ($\ell{=}25$) with the early-layer ($\ell{=}0$) features used in {\ourmethod}.
As shown in Table~\ref{tab:baseline_layer}, Cosine Uniqueness performs better with last-layer features, in contrast to {\ourmethod}, which benefits from early-layer features.
This indicates that the use of early-layer features is specific to {\ourmethod}, where smooth temporal trajectories directly translate to lower-residual prediction.

\subsection{Motion-Based Analysis} \begin{wraptable}{r}{0.3\columnwidth}
\centering
\captionsetup{width=0.3\columnwidth}
\caption{\scriptsize Performance on Video-MME for LLaVA-OneVision on RAFT Motion Brackets.}
\label{tab:llava-speed}
\scriptsize
\setlength{\tabcolsep}{4pt} 
\renewcommand{\arraystretch}{1.00}
\begin{tabular}{lcc}
\toprule
\textbf{Speed} & \textbf{Uniform} & \textbf{\ourmethod} \\
\midrule
Slow   & 57.7 & 59.0 {\tiny $\mathbf{\textcolor{ForestGreen}{(+1.3)}}$} \\
Medium & 59.1 & 58.9 {\tiny $\mathbf{\textcolor{Bittersweet}{(-0.2)}}$} \\
Fast   & 62.8 & 60.7 {\tiny $\mathbf{\textcolor{Bittersweet}{(-2.1)}}$} \\
\bottomrule
\end{tabular}
\vspace{0.2in}
\end{wraptable}
On Video-MME, we stratify results by motion intensity, using RAFT~\cite{teed2020raft} to compute global average optical flow and grouping videos into \textit{slow}, \textit{medium}, and \textit{fast} categories via $k$-means clustering (Table~\ref{tab:llava-speed}). With LLaVA-OneVision, \ourmethod\ yields the largest gain on \textit{slow} videos ($+1.3\%$), where high temporal redundancy provides ample room to discard uninformative frames. Performance remains neutral on \textit{medium} videos ($-0.2\%$) and regresses on \textit{fast} videos ($-2.1\%$). We hypothesize that in high-motion scenarios, consecutive frames differ substantially enough that uniform sampling provides diverse coverage and {\ourmethod}'s way of identifying temporal surprises might not yield sufficient temporal coverage.

\begin{craddedblock}
\subsection{Noise Sensitivity}
\label{app:cr_corruption}

Higher-order finite differences can amplify high-frequency perturbations. Although Swift Sampling does not include an explicit denoising module, it computes residuals from globally mean-pooled early-layer key features, providing implicit spatial smoothing before finite differences are applied.

To directly evaluate robustness, we apply frame-wise camera shake (random translation within $\pm5.0\%$ of the frame size), 11-pixel motion blur, and additive Gaussian noise ($\sigma=0.05$) to all candidate frames on LongVideoBench. We use the corrupted frames both for selection and as VLM input, and compare the default Swift Sampling configuration ($N=3$) with Uniform sampling under identical corruptions.

\begin{table}[htbp]
  \centering
  \ifcrreview\color{CRAdded}\captionsetup{font+={color=CRAdded},labelfont+={color=CRAdded}}\fi
  \small
  \setlength{\tabcolsep}{4pt}
  \caption{Accuracy under visual corruptions on LongVideoBench. Parentheses indicate the change from each method's clean-input accuracy.}
  \label{tab:cr_corruption}
  \begin{tabular}{lrrrr}
    \toprule
    Method & Clean & Camera shake & Motion blur & Gaussian noise \\
    \midrule
    Uniform & 55.3 & 55.2 ($-0.1$) & 54.3 ($-1.0$) & 54.2 ($-1.1$) \\
    Swift Sampling & \textbf{57.9} & \textbf{57.8} ($-0.1$) & \textbf{56.8} ($-1.1$) & \textbf{56.7} ($-1.2$) \\
    \bottomrule
  \end{tabular}
\end{table}

Across all three corruptions, Swift Sampling shows degradation comparable to Uniform sampling while retaining a 2.5--2.6-point advantage. Even under the most challenging setting, Gaussian noise, it achieves 56.7 accuracy compared with 54.2 for Uniform, confirming that its performance gain remains robust to visual corruption.
\end{craddedblock}

\begin{craddedblock}
\subsection{Analysis of Selected Frames}
\label{app:cr_selected_frames}

Swift Sampling does not explicitly distinguish semantic actions from shot transitions, as both can produce a large residual. We analyze the eight frames selected by Swift Sampling in Figure~\ref{fig:qual_drawing}, where the question asks for the temporal order of painting stages. We use an HSV color-histogram shot-boundary detector~\cite{zhang1993automatic} to identify shot transitions and manual annotations to assess each selected frame's relevance to the question (Table~\ref{tab:cr_frame_breakdown}).

\begin{table}[htbp]
  \centering
  \ifcrreview\color{CRAdded}\captionsetup{font+={color=CRAdded},labelfont+={color=CRAdded}}\fi
  \small
  \caption{Frame-level breakdown of the eight frames selected by Swift Sampling in Figure~\ref{fig:qual_drawing} for the question about the temporal order of painting stages.}\vspace{0.1in}
  \label{tab:cr_frame_breakdown}
  \begin{tabular}{lr}
    \toprule
    Category & Selected frames \\
    \midrule
    Question-relevant within-shot painting progression & 4/8 \\
    Question-relevant painting stage at a framing cut & 1/8 \\
    Other selected frames & 3/8 \\
    \bottomrule
  \end{tabular}
\end{table}

Overall, 5/8 selected frames directly support the correct event ordering $1\rightarrow4\rightarrow2\rightarrow3$. Four capture relevant painting progression within a shot, while one captures a relevant stage at a framing transition. In this example, Swift Sampling selects both a task-relevant shot transition and task-relevant within-shot changes.
\end{craddedblock}

\section{Qualitative analysis}\label{app:qualitative}
\subsection{Additional Results}
In this section, we provide additional qualitative examples across different datasets, video types, and frame budgets.

\begin{figure*}[t]
\centering
\includegraphics[width=\textwidth]{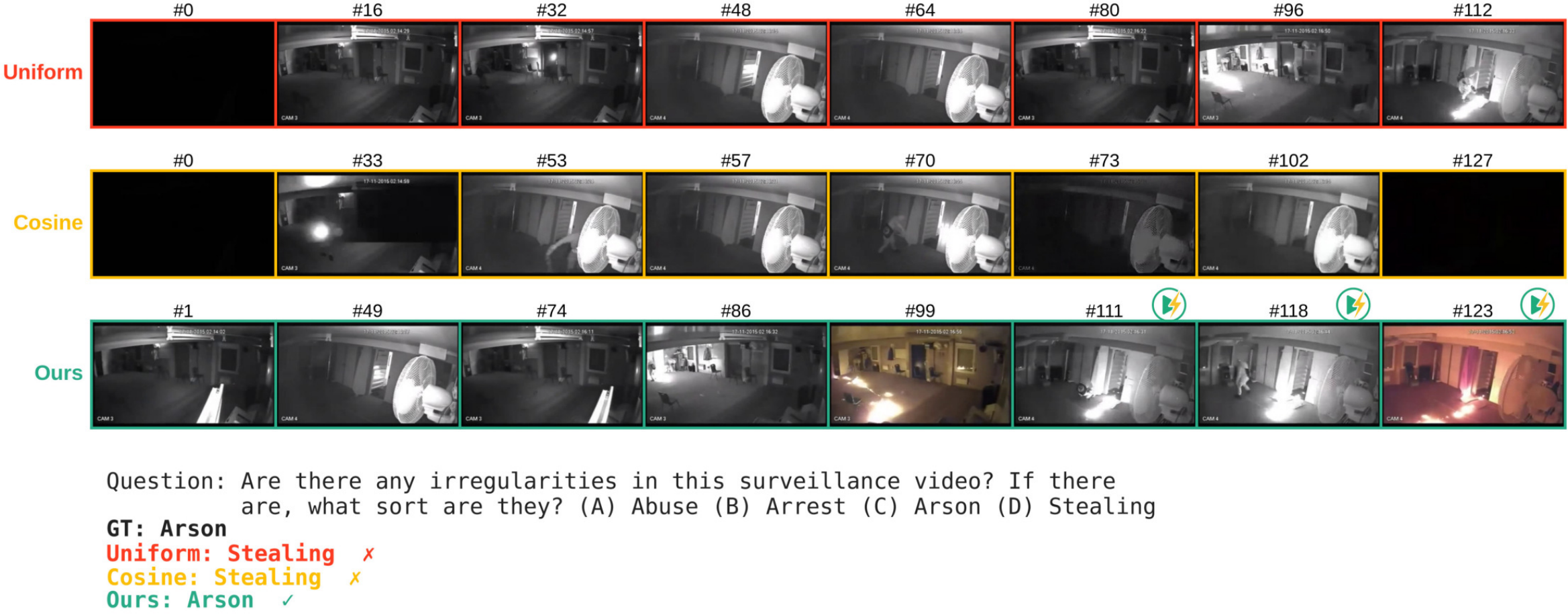}
\caption{\footnotesize \textbf{Additional qualitative example} on a MLVU surveillance video, selecting $8$ frames out of $128$. The correct answer requires identifying a fire breakout late in the video. Uniform sampling distributes frames evenly and captures only a brief glimpse of the fire. Cosine Uniqueness~\cite{yuan2025unicomp} concentrates on visually distinctive but content-redundant frames in the dark surveillance footage and entirely misses the fire. Our method captures the transition from normal surveillance to fire breakout, providing the video LLM with the temporal evidence needed to correctly identify the irregularity as arson. Crucially, {\ourmethod} successfully captures \textbf{local changes, e.g., a person entering between frame 111 and frame 118} as well as \textbf{global changes such as fire breakout (frame 123)}.}
\label{fig:qual_arson}
\end{figure*}

\begin{figure*}[t]
\centering
\includegraphics[width=0.9\textwidth]{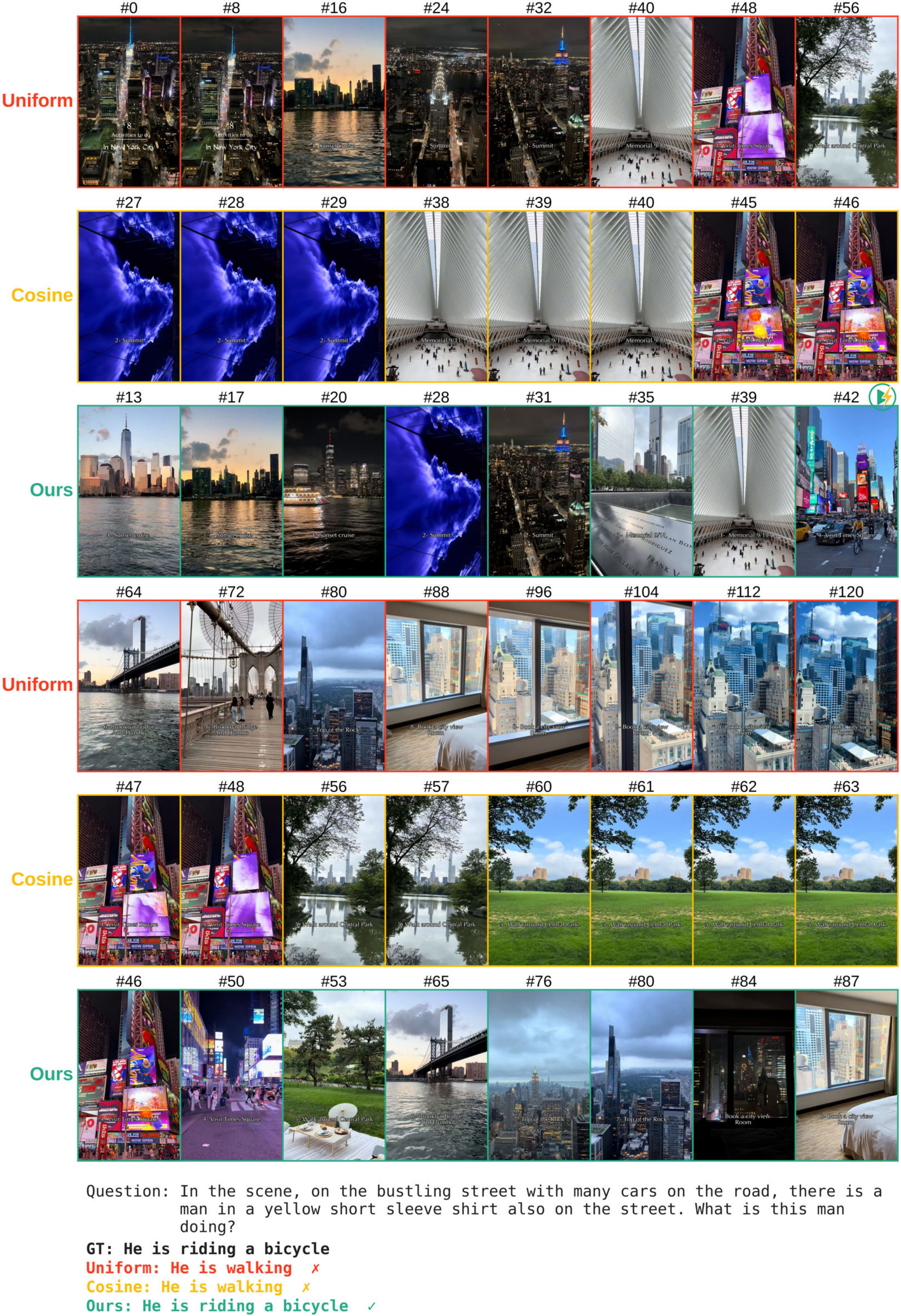}
\caption{\footnotesize \textbf{Qualitative example from the LVB dataset.}
Each row shows the $16$ frames selected by Uniform sampling (\textcolor{red}{red}), Cosine Uniqueness~\cite{yuan2025unicomp} (\textcolor{orange}{yellow}), and our method (\textcolor{green!50!black}{green}) from the full $128$-frame candidate pool. The question asks what the man in a yellow short-sleeve shirt is doing in a bustling street scene, with the correct answer being \textit{riding a bicycle}.
Answering this question requires retaining the specific street scene described in the question and the local action within it.
Uniform sampling and Cosine Uniqueness both select visually diverse city scenes, but they miss the target moment around frame \#42 and therefore predict \textit{walking}.
In contrast, our method selects a more diverse set of frames across the video while still retaining the question-relevant street scene around frame \#42. As a result, it preserves the frame needed to correctly identify that the man is riding a bicycle. 
}
\label{fig:qual_lvb_bicycle}
\end{figure*}

\begin{figure*}[t]
\centering
\includegraphics[width=\textwidth]{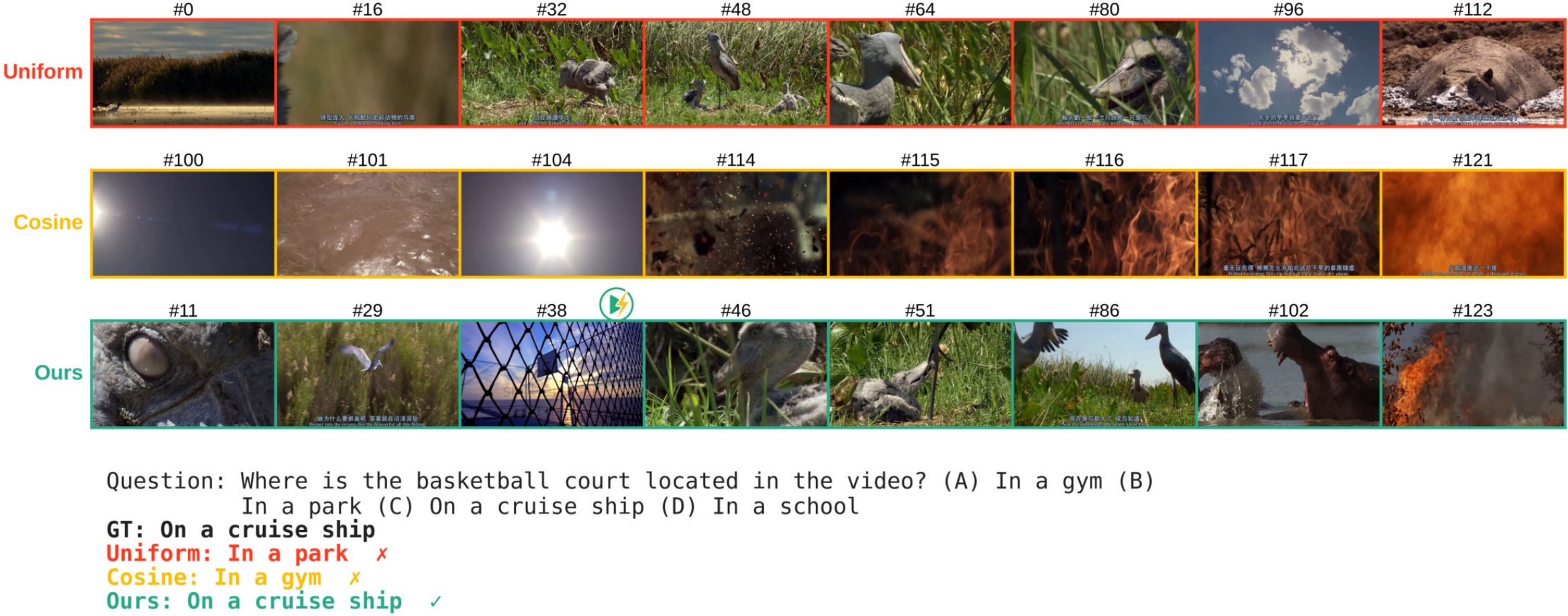}
\caption{\footnotesize \textbf{Qualitative example from the MLVU dataset.}
Each row shows the $8$ frames selected by Uniform sampling (\textcolor{red}{red}), Cosine Uniqueness~\cite{yuan2025unicomp} (\textcolor{orange}{yellow}), and our method (\textcolor{green!50!black}{green}) from the full $128$-frame candidate pool.
The question asks where the basketball court in the video is located, with the correct answer being \textit{on a cruise ship}.
The video contains highly diverse scenes, so answering the question requires retaining broad scene coverage together with the brief contextual cues that indicate the cruise-ship setting.
Uniform sampling captures mainly natural scenes and also includes visually similar selections, leading the model to predict \textit{in a park}.
Cosine Uniqueness concentrates on a narrow subset of visually distinctive frames, such as flames and bright sky shots, leaving the cruise-ship context under-covered.
In contrast, our method selects a more diverse set of frames across the video and captures the basketball-court scene in frame \#38.
In this frame, the court appears together with the surrounding waterway, allowing the model to infer that it is located on a cruise ship.
As a result, it correctly answers the question. 
}
\label{fig:qual_mlvu_cruise}
\end{figure*}

\begin{figure*}[t]
\centering
\includegraphics[width=\textwidth]{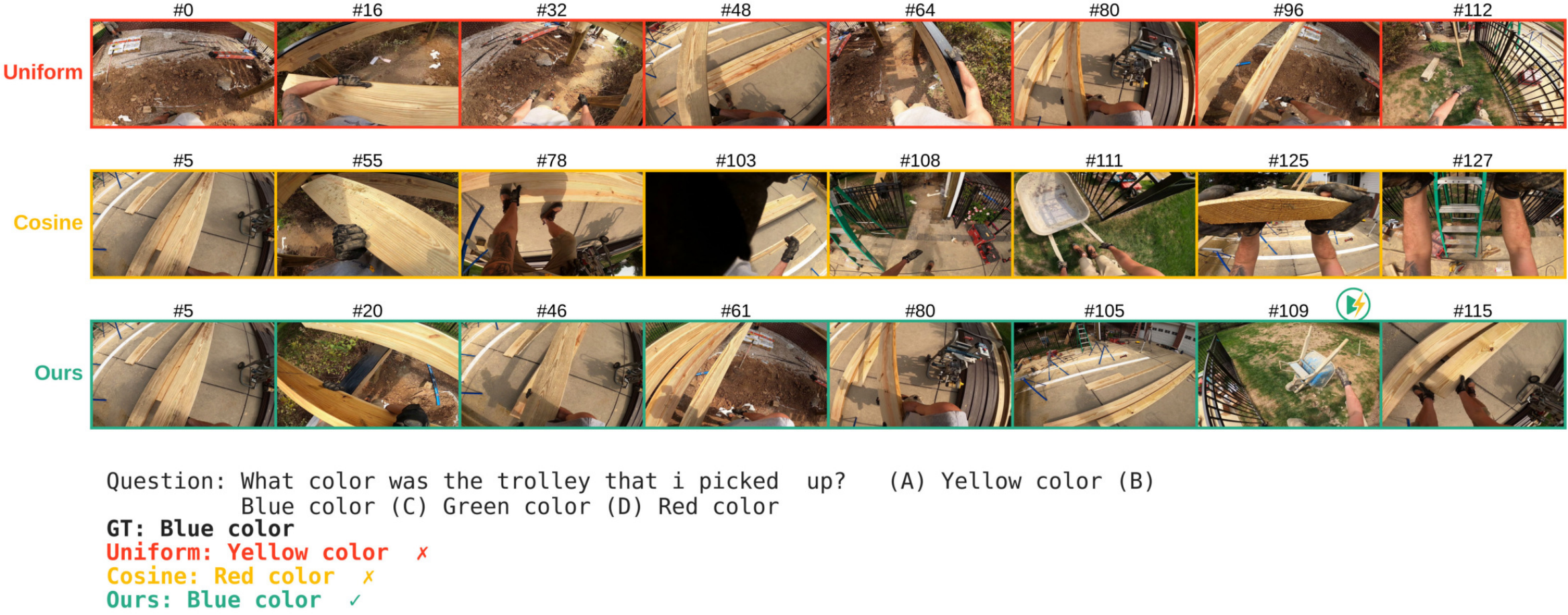}
\caption{\footnotesize \textbf{Qualitative example from the MLVU dataset.}
Each row shows the $8$ frames selected by Uniform sampling (\textcolor{red}{red}), Cosine Uniqueness~\cite{yuan2025unicomp} (\textcolor{orange}{yellow}), and our method (\textcolor{green!50!black}{green}) from the full $128$-frame candidate pool.
The question asks for the color of the trolley picked up in the video, with the correct answer being \textit{blue}.
This is an egocentric video of a construction-related task, and answering the question requires capturing the brief moment when the trolley is picked up.
Uniform sampling misses this moment entirely.
Cosine Uniqueness captures the trolley, but only partially, making its color difficult to identify.
In contrast, our method captures the pickup moment with the trolley clearly visible, preserving the information needed to correctly identify its color. }
\label{fig:qual_mlvu_trolley}
\end{figure*}

\begin{figure*}[t]
\centering
\includegraphics[width=\textwidth]{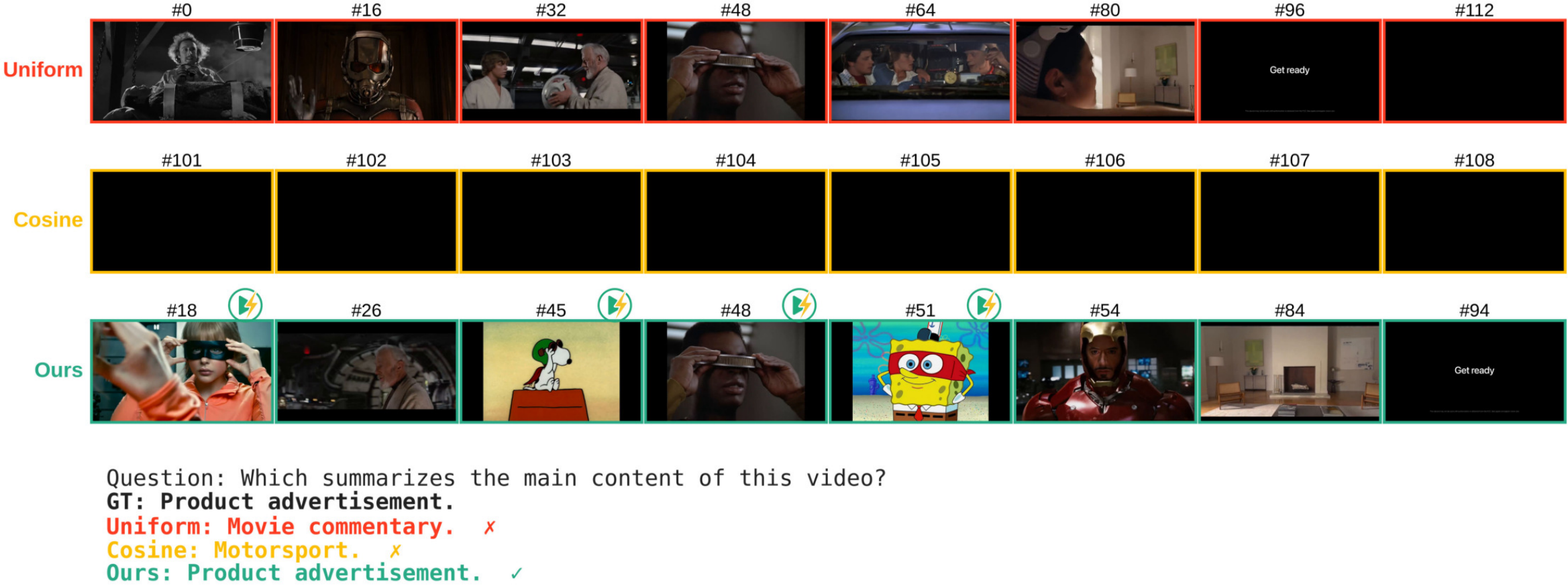}
\caption{\footnotesize \textbf{Qualitative example from the Video-MME dataset under a tight frame budget.}
Each row shows the $8$ frames selected by Uniform sampling (\textcolor{red}{red}), Cosine Uniqueness~\cite{yuan2025unicomp} (\textcolor{orange}{yellow}), and our method (\textcolor{green!50!black}{green}) from the full $128$-frame candidate pool.
The question asks which option best summarizes the main content of the video, with the correct answer being \textit{Product advertisement}.
This question requires integrating evidence across multiple scenes rather than recognizing a single salient moment.
Uniform sampling captures a range of scenes, but because the selected frames are spaced evenly, they resemble a sequence of unrelated movie shots and lead the model to predict \textit{Movie commentary}.
Cosine Uniqueness concentrates on highly distinctive black transition frames under this limited frame budget, leaving too few informative frames to recover the video's overall theme.
In contrast, our method selects multiple device-centric frames along with diverse contextual scenes, better preserving the recurring visual cues that characterize the video as a product advertisement. As a result, it preserves the information needed to correctly identify the video as a product advertisement.}
\label{fig:qual_product_ad}
\end{figure*}

\begin{figure*}[t]
\centering
\includegraphics[width=\textwidth]{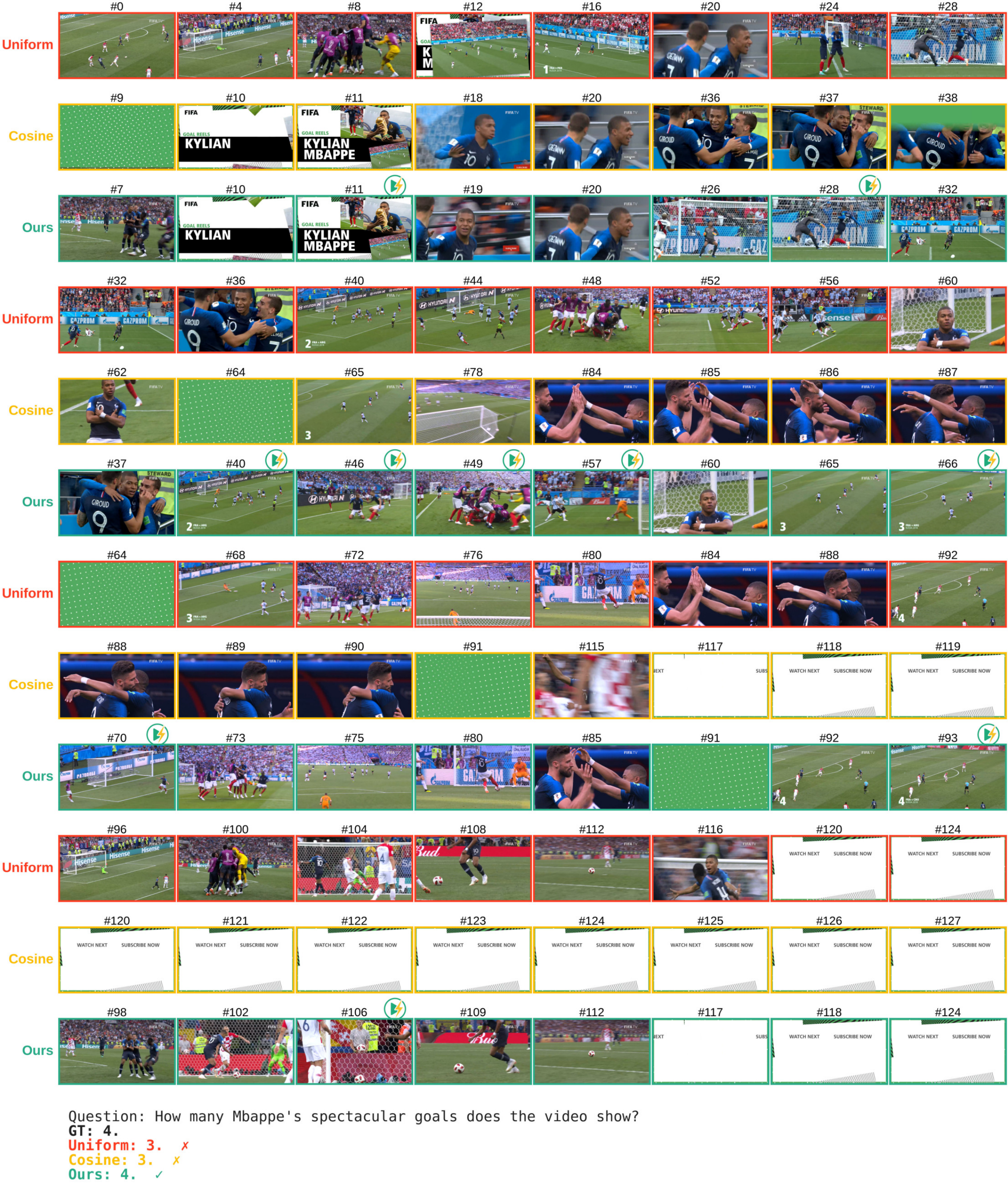}
\caption{\footnotesize \textbf{Qualitative example from the Video-MME dataset.}
Each row shows the $32$ frames selected by Uniform sampling (\textcolor{red}{red}), Cosine Uniqueness~\cite{yuan2025unicomp} (\textcolor{orange}{yellow}), and our method (\textcolor{green!50!black}{green}) from the full $128$-frame candidate pool.
The question asks how many of Mbapp\'e's spectacular goals are shown in the video, with the correct answer being $4$.
The video shows four separate goal segments, each marked by a numbered overlay.
Uniform sampling distributes frames evenly but misses one of these goal segments.
Cosine Uniqueness selects visually salient frames, such as close-up shots and repeated graphic frames, but does not cover all of the numbered goal segments needed for counting.
In contrast, our method selects a more diverse set of frames across the video with good temporal coverage, capturing frames corresponding to goals \#1, \#2, \#3, and \#4.
As a result, it retains the frames needed to correctly answer the counting question.}
\label{fig:qual_mbappe_count}
\end{figure*}

\begin{figure*}[t]
\centering
\includegraphics[height=0.7\textheight]{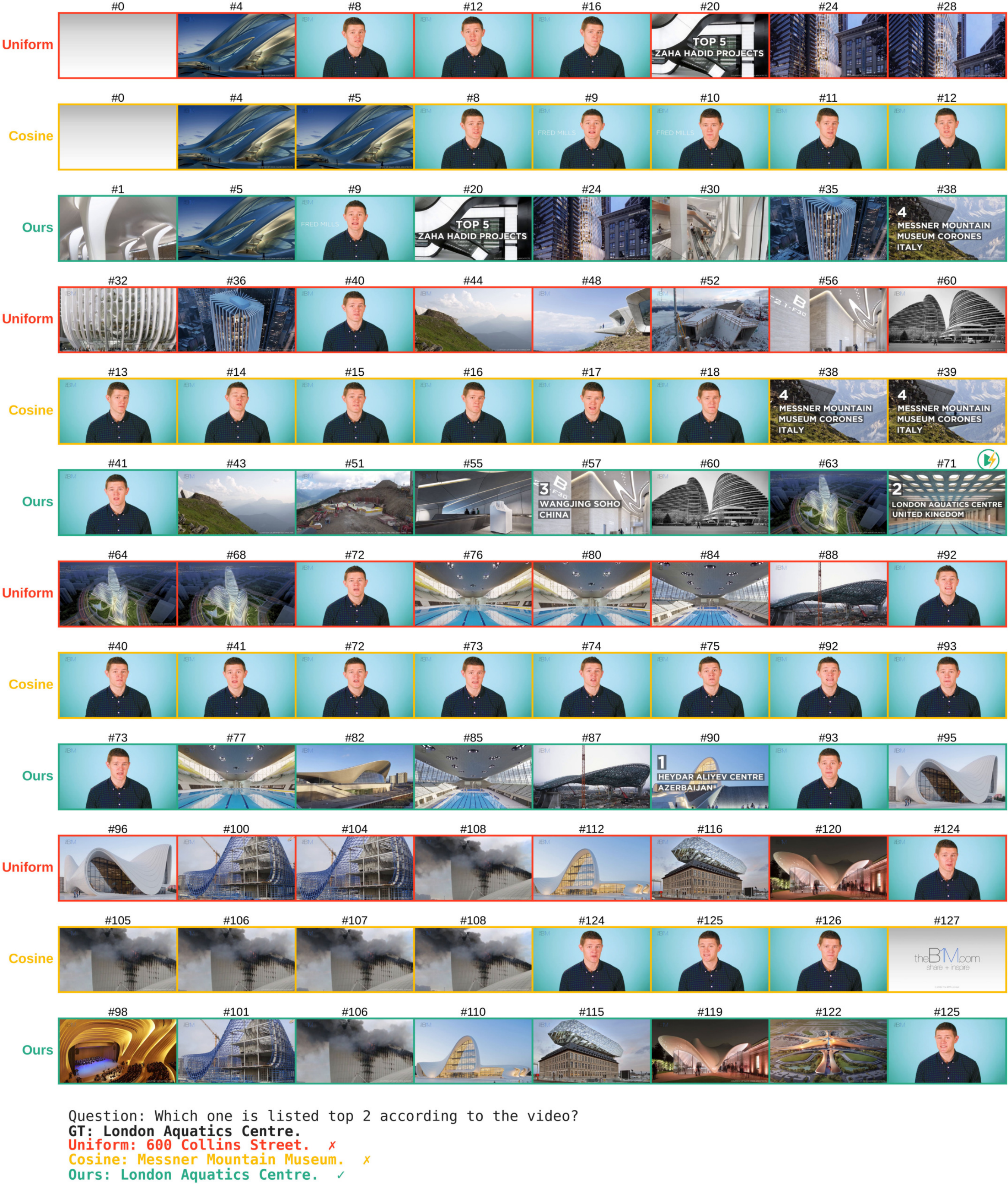}
\caption{\footnotesize \textbf{Qualitative example on an informational video} from the Video-MME dataset.
Each row shows the $32$ frames selected by Uniform sampling (\textcolor{red}{red}), Cosine Uniqueness~\cite{yuan2025unicomp} (\textcolor{orange}{yellow}), and our method (\textcolor{green!50!black}{green}) from the full $128$-frame candidate pool.
The question asks which item is listed in the top 2, with the correct answer being \textit{London Aquatics Centre}.
Since the video presents several places one after another, the model needs frames that cover the relevant places and their order in the video. Uniform sampling provides broad temporal coverage but still contains redundant selections.
Cosine Uniqueness captures visually distinctive frames, but several of its selected frames come from similar moments of the same entries, so it misses some of the places needed to answer the question. 
In contrast, our method selects a more diverse set of frames across the video, providing broader coverage of the entries and retaining the question-relevant moments (\# 71). As a result, it preserves the evidence needed to correctly identify the top-2 item.}
\label{fig:qual_informational_top2}
\end{figure*}

\paragraph{Frame selection examples.}
\Cref{fig:qual_lvb_bicycle,fig:qual_mlvu_cruise,fig:qual_mlvu_trolley,fig:qual_product_ad,fig:qual_mbappe_count,fig:qual_informational_top2,fig:qual_arson} compare the frames selected by Uniform sampling, Cosine Uniqueness~\cite{yuan2025unicomp}, and Swift Sampling across Video-MME, LongVideoBench, and MLVU.
These examples illustrate that Swift Sampling provides broader temporal coverage while reducing redundant selections, which helps preserve frames needed for different types of long-video understanding.

\begin{figure*}[t]
\centering
\includegraphics[height=0.85\textheight]{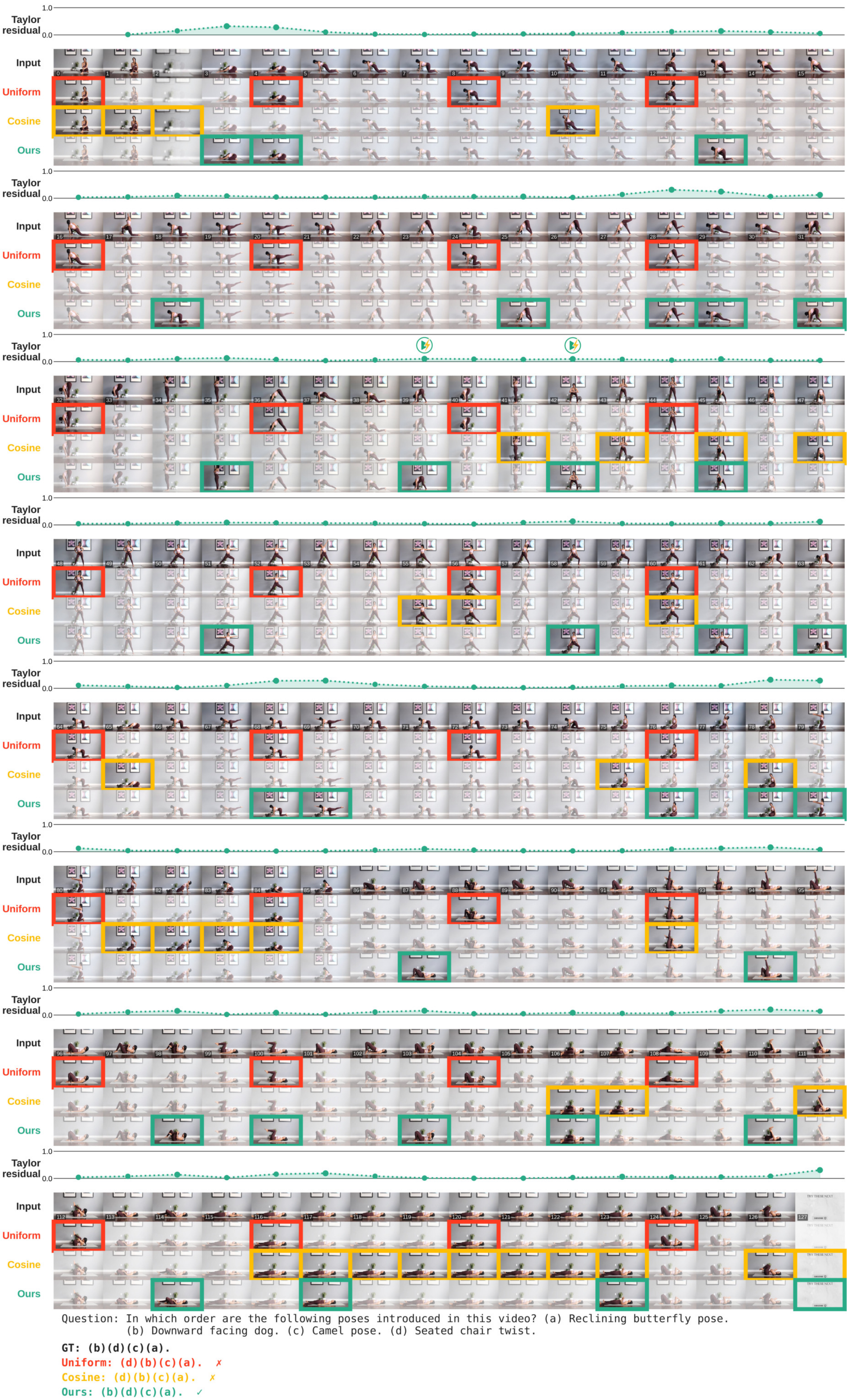}
\caption{\footnotesize \textbf{Qualitative example on a visually static yoga video from the Video-MME dataset.}
Each block shows the Taylor residual over time (\textit{top}) and the corresponding input frames together with the $32$ frames selected by Uniform sampling (\textcolor{red}{red}), Cosine Uniqueness~\cite{yuan2025unicomp} (\textcolor{orange}{yellow}), and our method (\textcolor{green!50!black}{green}) from the full $128$-frame candidate pool.
The question asks for the order in which four poses are introduced, with the correct answer being (b)\,(d)\,(c)\,(a).
Unlike videos with frequent scene cuts, this clip remains visually static, so the relevant evidence lies in subtle pose transitions.
Uniform and Cosine Uniqueness both miss the seated chair twist around frame \#42 and therefore confuse (b) downward facing dog with (d) seated chair twist.
Our method captures this key transition and correctly answers the question, illustrating that temporal surprise can identify informative moments even when motion is small and the overall scene changes very little.}
\label{fig:qual_yoga_order}
\end{figure*}

\begin{figure*}[t]
\centering
\includegraphics[height=0.85\textheight]{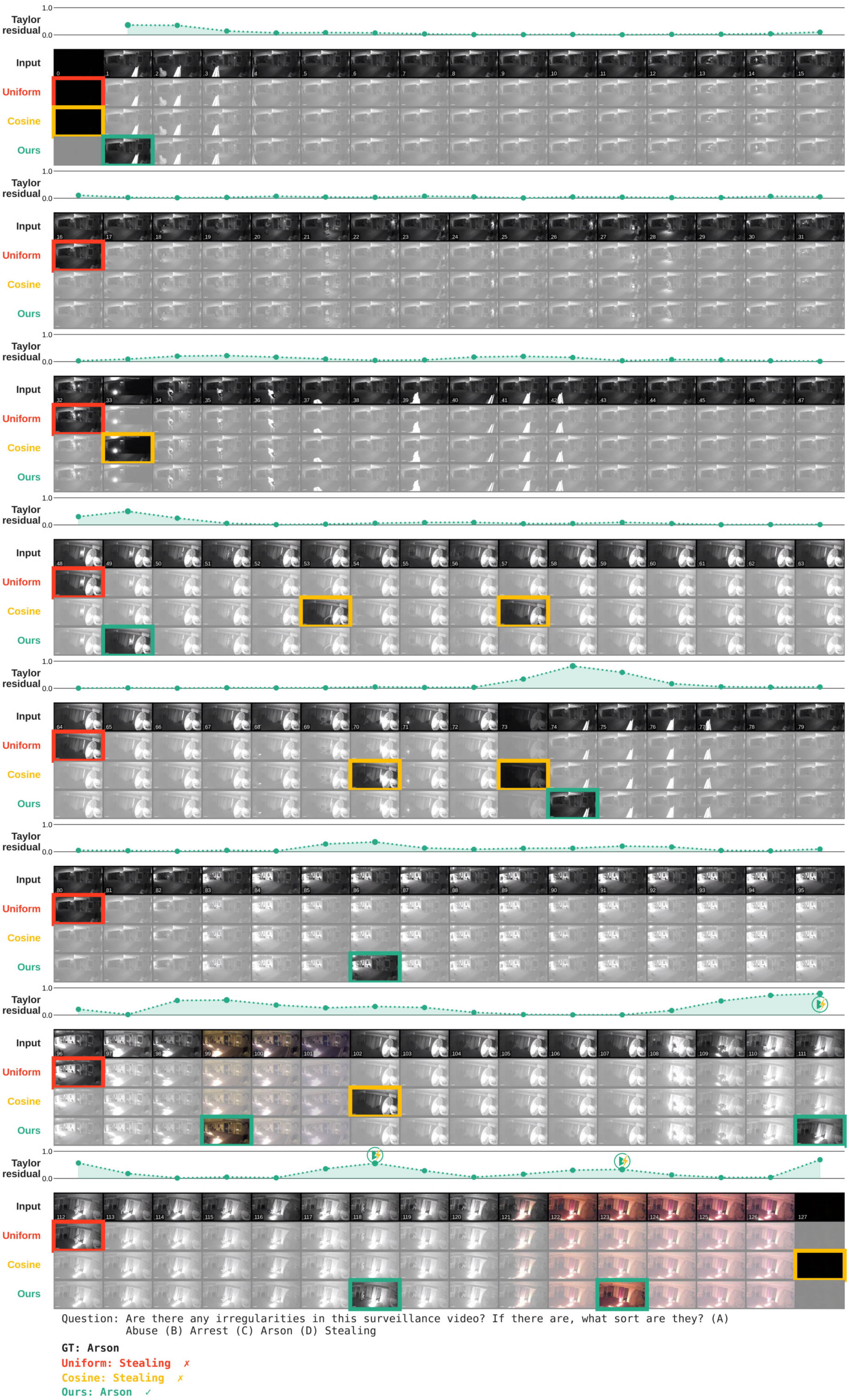}
\caption{\footnotesize \textbf{Qualitative example from the MLVU dataset.}
Each block shows the Taylor residual over time (\textit{top}) and the corresponding input frames together with the $8$ frames selected by Uniform sampling (\textcolor{red}{red}), Cosine Uniqueness~\cite{yuan2025unicomp} (\textcolor{orange}{yellow}), and our method (\textcolor{green!50!black}{green}) from the full $128$-frame candidate pool.
The question asks whether there is any irregularity in the surveillance video and, if so, what type it is, with the correct answer being \textit{arson}.
The video is largely composed of repetitive grayscale surveillance footage, and the key evidence appears only when the scene changes into a fire breakout near the end.
Uniform sampling distributes frames evenly across the clip and captures only a weak glimpse of this anomaly, leading the model to predict \textit{stealing}.
Cosine Uniqueness concentrates on visually distinctive but content-redundant frames in the surveillance footage and fails to preserve the fire breakout.
In contrast, our method captures the transition from normal surveillance to the arson scene, preserving the anomaly-related frames needed to correctly identify the irregularity as \textit{arson}.}
\label{fig:qual_arson_full}
\end{figure*}

\paragraph{Taylor residual visualizations.}
\Cref{fig:qual_yoga_order,fig:qual_arson_full,fig:app_teaser}
show examples with Taylor residual scores plotted above the frame sequence.
Across these cases, high-residual frames often align with informative temporal changes, such as subtle pose transitions, scene changes, and anomaly events.
These examples support our intuition that temporal surprise is an effective signal for selecting informative frames in long videos.


\begin{figure*}[t]
\centering
\includegraphics[width=\textwidth]{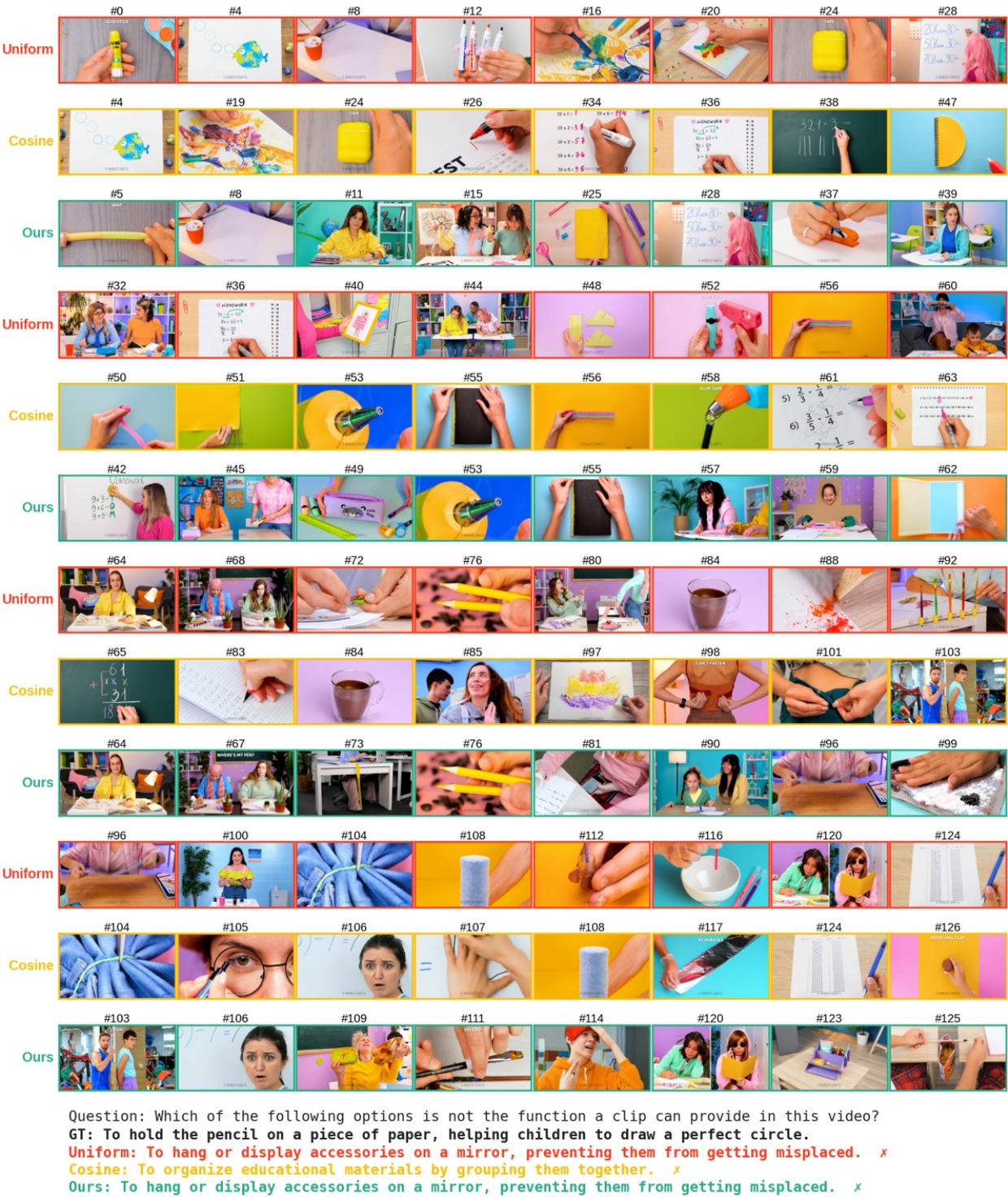}
\caption{\footnotesize \textbf{Failure case} from the Video-MME dataset, selecting $32$ frames out of $128$. This video is composed of highly diverse, rapidly changing scenes, with almost every segment corresponding to a new scene cut. Consequently, the advantage of sophisticated frame selection is reduced: even Uniform sampling already captures broadly distinct frames. The question asks which function of a clip is \emph{not} shown in the video; the correct answer is using a clip to hold a pencil on paper to help draw a perfect circle. Because the video contains many unrelated clip-based demonstrations and the relevant evidence is brief, all methods are misled by other plausible clip usages. Uniform sampling and our method predict hanging accessories on a mirror, while Cosine Uniqueness predicts organizing educational materials. This example shows that when a video has little temporal redundancy, \textbf{all frame selection methods} offer limited gains.}
\label{fig:failure}
\end{figure*}

\subsection{Failure Cases}

\begin{figure*}[t]
\centering
\includegraphics[width=\textwidth]{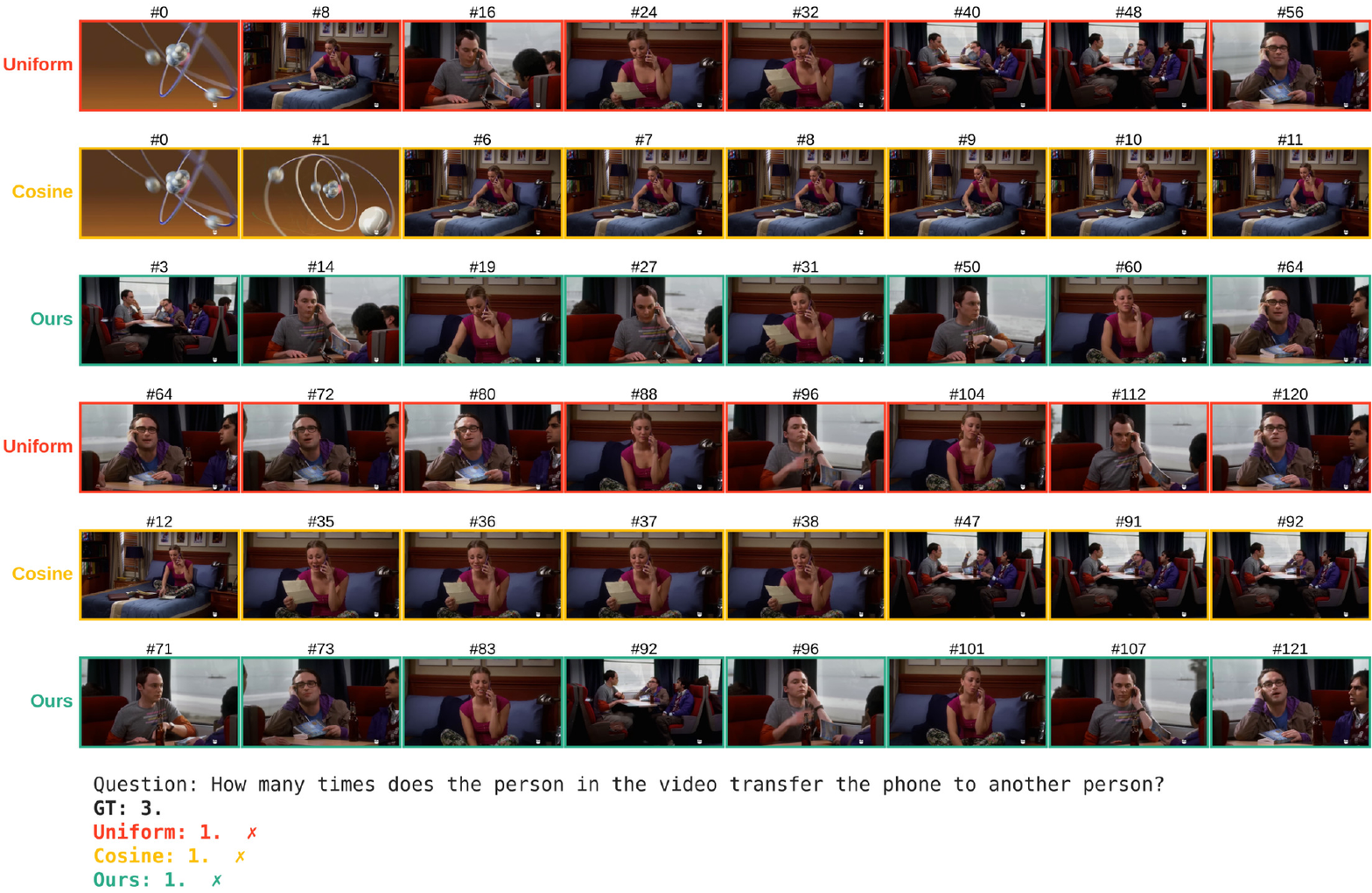}
\caption{\footnotesize \textbf{Failure case: query-specific reasoning under tight frame budgets.}
On a Video-MME counting question (``How many times does the person transfer the phone to another person?''; GT: $3$), all three methods select $16$ frames from $128$ candidates and incorrectly answer $1$.
Uniform sampling distributes frames evenly across the clip but captures only the static phone-call segments between transfers.
Cosine Uniqueness clusters around visually distinct frames such as the opening title card, leaving the relevant action under-sampled.
While our method covers diverse character views across the clip, only $2$ phone-transfer events are observable in the selected frames.
Without explicit query conditioning, visually informative frames do not always coincide with the frames most relevant to the question.
}
\label{fig:qual_phone}
\end{figure*}
Swift Sampling can fail when the question depends on frames that are not selected by a query-agnostic criterion.
In Fig.~\ref{fig:qual_phone}, the question asks how many times one person hands the phone to another person, with the correct answer being $3$.
Our method covers diverse character views, but misses some phone-transfer moments under the $16$-frame budget, causing the model to answer $1$.
We also observe limited benefits when the video has rapid scene changes and little temporal redundancy, as shown in Fig.~\ref{fig:failure}.
In this case, most candidate frames are already visually distinct, so Uniform sampling can already select diverse frames.
In such cases, adding query-aware signals can further improve performance, at the cost of efficiency, as shown in Table~\ref{tab:main_results}.

\subsection{Details of Fig.~\ref{fig:teaser}}
\begin{figure*}[t]
\centering
\includegraphics[height=0.9\textheight]{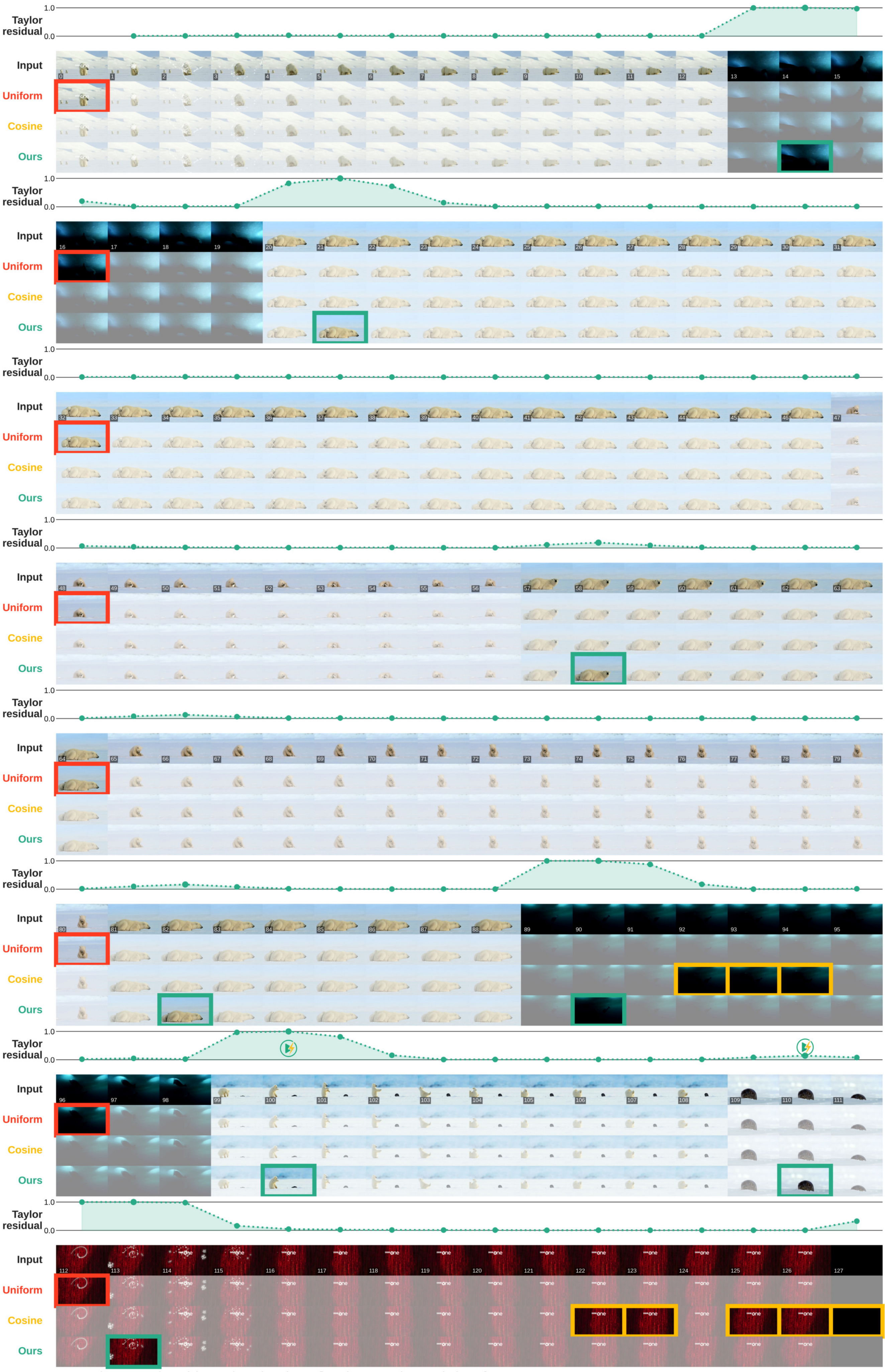}
\caption{\footnotesize \textbf{Detailed view of Fig.~\ref{fig:teaser}.} Full frame selection results on the same video.
Each block shows the Taylor residual over time (\textit{top}) and the corresponding input frames together with the $32$ frames selected by Uniform sampling (\textcolor{red}{red}), Cosine Uniqueness~\cite{yuan2025unicomp} (\textcolor{orange}{yellow}), and our method (\textcolor{green!50!black}{green}) from the full $128$-frame candidate pool.
Our method selects temporally surprising moments. In contrast, uniform sampling distributes frames evenly, sometimes resulting in redundant selections. Cosine Uniqueness sometimes concentrates on a narrow set of visually distinctive but content-redundant frames.}
\label{fig:app_teaser}
\end{figure*}

In Fig.~\ref{fig:teaser}, we visualize a partial view of the frame selection results from three methods: uniform sampling, Cosine Uniqueness, and our method, selecting $8$ frames out of $128$ candidates. Here we present the full selection results for this video.
As shown in Fig.~\ref{fig:app_teaser}, the Taylor residual effectively reflects the informativeness of each frame, selecting the most ``surprising'' frames where the visual feature trajectory changes sharply.
In contrast, uniform sampling often selects redundant frames, wasting the frame budget.
Cosine Uniqueness selects visually unique frames, but as it scores frames by their global visual distinctiveness within the video, it tends to concentrate on a narrow set of distinctive but content-redundant frames, as also illustrated in Fig.~\ref{fig:qual_example}.

\subsection{Task Category Examples}
\label{app:qual_task_analysis}
\begin{figure*}[t]
\centering
\includegraphics[width=\textwidth]
{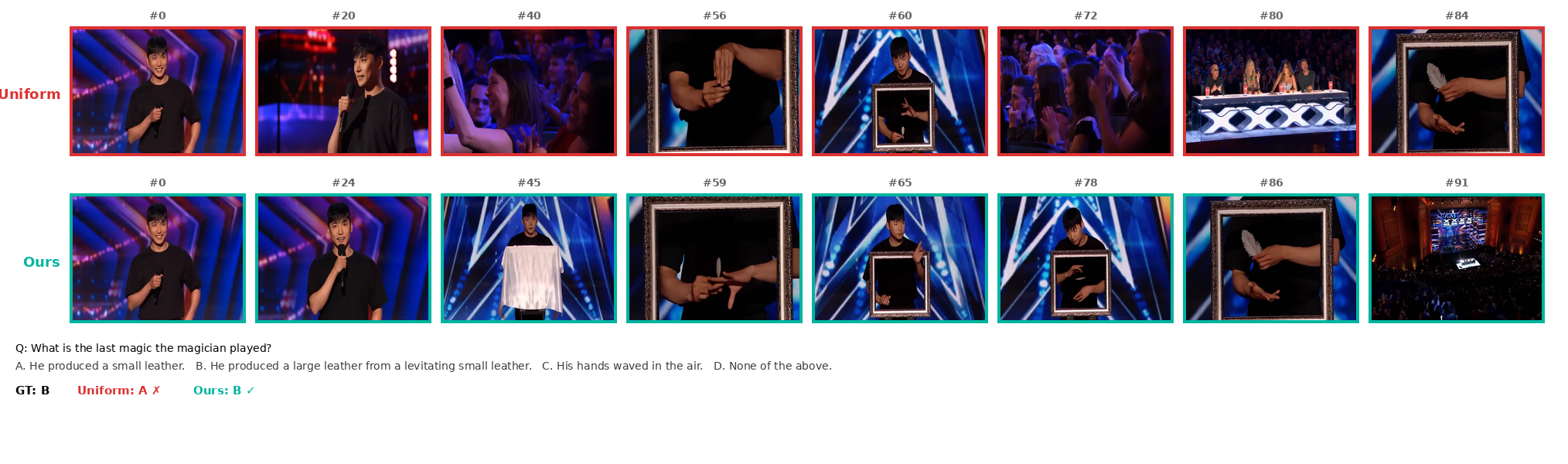}
\caption{\footnotesize \textbf{Qualitative Example: Action Recognition} from the Video-MME dataset. The question asks \textit{``What is the
last magic the magician played?''}. 8 Key frames corresponding the last magic trick are being showcased for Uniform sampling and our method.  Correctly identifying the action requires capturing 3 key moments: the introduction of the small feather, the small feather levitating and the appearance of the large feather. Our method is able to select all three frames (frame \#59, \#78 and \#86) while Uniform sampling misses the levitation making the transition from small feather to large feather ambiguous}
\label{fig:qual_magician}
\end{figure*}
\Cref{fig:qual_magician} compares frames selected by Uniform Sampling and \ourmethod~on a question from the \textit{Action Recognition} task in the Video-MME dataset. This example showcases how \ourmethod~is able to select key frames representative of the motion in the video whereas Uniform sampling can miss brief but decisive moments.
\begin{figure*}[t]
\centering
\includegraphics[width=\textwidth]
{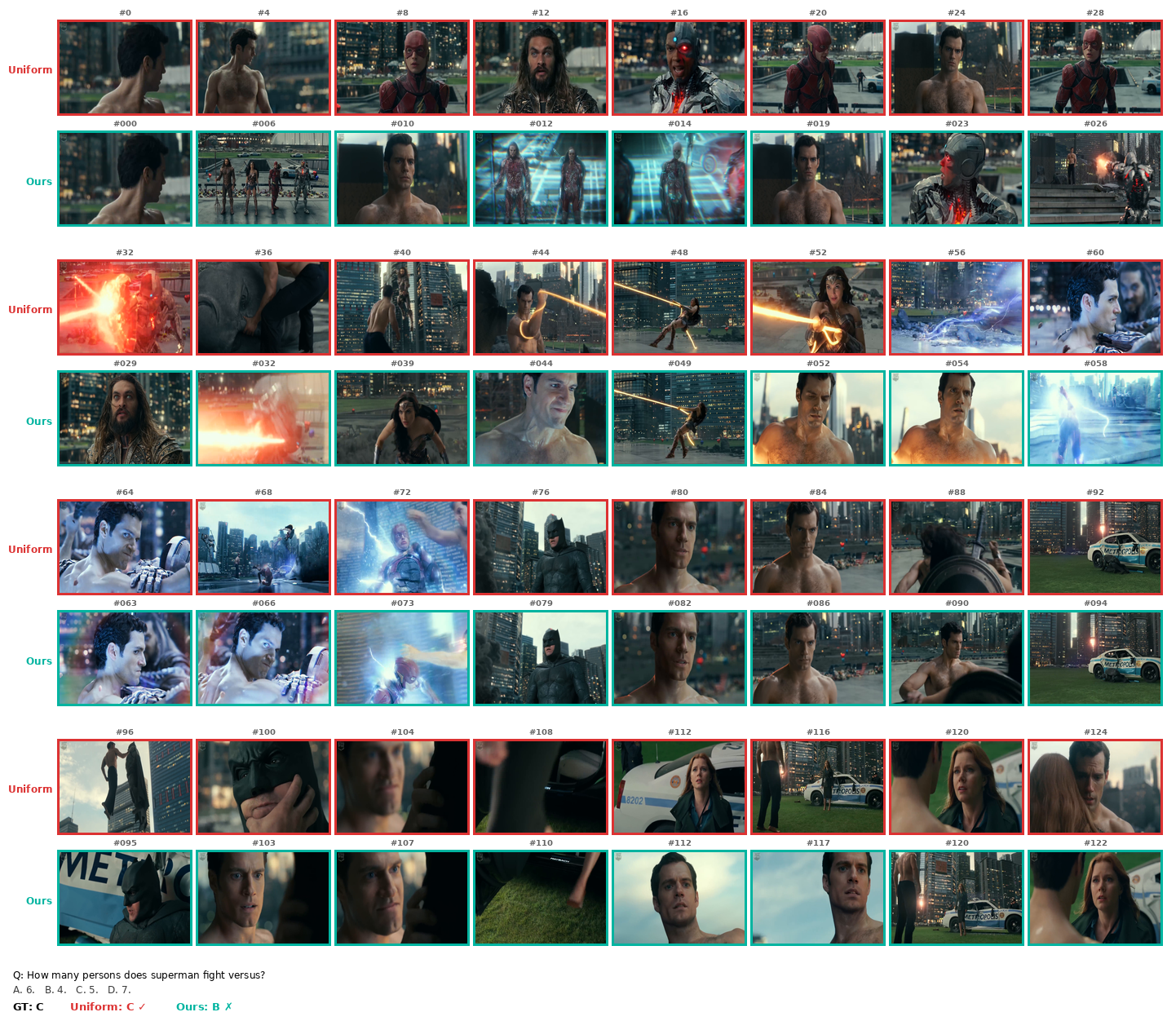}
\caption{\footnotesize \textbf{Qualitative Example: Counting} from the Video-MME dataset. The question asks \textit{``How many persons does superman fight versus?''}. All 32 frames selected by Uniform sampling and our method have been showcased. Uniform sampling includes frames like frame \#80 where Superman holding Batman mid-air which are missed by our method. Without such frames, the model likely fails to recognize Batman as a distinct opponent, under-counting by one (4 instead of 5).}
\label{fig:qual_superman_counting}
\end{figure*}
\Cref{fig:qual_superman_counting} illustrates a failing example of \ourmethod~on a \textit{Counting} question in the Video-MME dataset. The video depicts Superman fighting five opponents across different scenes. Uniform sampling, by evenly spanning the timeline, captures each combatant at least once in an engagement with Superman. \ourmethod~has coverage gaps by missing scenes that also identify Batman as a combatant. Without such visual evidence, the model underestimates the answer (4 instead of 5). This exemplifies how counting tasks rely on exhaustive temporal coverage, where even visually unremarkable moments carry task-relevant information.

\section{Broader Impacts}\label{sec:broader_impacts}
Our work presents a training-free frame selection method to improve the performance of Video Large Language Models (VLLMs) on long-video understanding. VLLM pipelines are generic technologies with a wide range of potential applications in areas such as education, accessibility, content summarization, and video search. We are unaware of all potential applications, but we are cognizant that each use case has its own societal impact depending on the intentions and motivations of the individuals or organizations deploying the system. For example, long-video understanding can be used for productive purposes, such as a user asking a system detailed questions about long instructional videos, but it may also be used for applications such as automated surveillance systems.

\clearpage

\end{document}